\documentclass[12pt]{article}
\usepackage[preprint]{neurips_2026}
\usepackage[T1]{fontenc}
\usepackage{microtype}
\usepackage{hyperref}
\usepackage{url}
\usepackage{multirow}
\usepackage{booktabs}
\usepackage{graphicx}
\usepackage{amsmath}
\usepackage{enumitem}
\usepackage[font=small,labelfont=bf]{caption}
\usepackage{placeins}
\usepackage{float}
\usepackage{adjustbox}
\setcitestyle{authoryear,round,comma}

\usepackage{lineno}
\usepackage{rotating}
\title{The Computational Basis of Confidence in Large Language Models}

\author{%
  Dharshan Kumaran\thanks{Corresponding author: \texttt{dkumaran@google.com}} \\
  Google DeepMind
  \And
  Viorica Patraucean \\
  Google DeepMind
  \And
  Maks Ovsjanikov \\
  Google DeepMind and \'Ecole Polytechnique  \AND
  Petar Veli\v{c}kovi\'{c} \\
  Google DeepMind
  \And
  Nathaniel Daw \\
  Princeton University, USA
}

\begin{document}
\maketitle
\section*{Abstract}
Reliable confidence—the probability that a model's own answer is correct—is essential for the trustworthy deployment of language models \citep{steyvers2025metacognition, xiong2023can, tian2023just, kadavath2022language}. Existing work has largely evaluated confidence according to how well it predicts correctness and whether it is calibrated, leaving open a more fundamental question: what does the confidence signal itself represent? Answer logits may reflect a latent decision variable sufficient to compute normative confidence, or instead a heuristic preference signal that combines the available evidence in a non-Bayesian manner. We address this question using statistical decision confidence (SDC), a normative framework from computational neuroscience \citep{kepecs2008neural, masset2020behavior, lak2014orbitofrontal, hangya2016mathematical, sanders2016signatures}. Treating the answer-logit difference (LD) as a candidate readout of the latent decision variable, we test the qualitative signatures predicted by SDC. Across three perceptual discrimination tasks and a memory-based decision task, spanning three multimodal non-reasoning language models and one multimodal reasoning model, LD satisfied these signatures—including the diagnostic correct/error folded-X pattern—demonstrating that, in these settings, answer logits behave as monotonic readouts of a latent decision variable rather than merely heuristic preference scores. This signal was also behaviourally consequential: when allowed to abstain, models withheld low-confidence answers in a near-optimal manner, raising the accuracy of committed responses to near ceiling. In complex visual reasoning, LD continued to predict correctness beyond objective task difficulty, but the full geometric signatures of SDC were no longer observed, illustrating the current boundary of the framework when explicit normative process models are unavailable. These results provide a computational account of confidence in multimodal language models, delineate the conditions under which answer logits behave as readouts of a latent decision variable rather than heuristic preference scores, and establish statistical decision confidence as a unifying computational framework for studying confidence across biological and artificial intelligence.

\section*{Introduction}
Confidence is usually understood as a system's estimate of whether its own answer is correct \citep{pouget2016confidence, fleming2017self, kepecs2012computational, mamassian2016visual}. For large language models (LLMs), several candidate confidence signals have been proposed, including answer-token probabilities, consistency across repeated samples, and direct verbal reports such as ``I am 80\% confident'' \citep{xiong2023can, kadavath2022language, steyvers2025metacognition, tian2023just}. Among these, answer logits are perhaps the most fundamental, as they directly determine which response the model selects. Existing work has largely evaluated such signals according to a single criterion---how well they predict correctness---and under this empirical definition, answer logits are widely regarded as confidence signals.

However, predictive accuracy alone does not establish what these signals represent. A large answer-logit difference indicates that the model strongly favours one answer over competing alternatives --- and specifies a probability distribution over options --- but the computation underlying this preference is largely unspecified. One possibility is that answer logits are a monotonic readout of a latent decision variable encoding noisy task-relevant evidence---a variable that, under a normative process model defined by the task, is sufficient in principle to compute the posterior probability that the current choice is correct (the posterior confidence). Alternatively, answer logits may instead reflect a heuristic score. This score may incorporate task-irrelevant information, or combine task-relevant information in a manner that deviates from an ideal observer, and so need not read out this decision variable at all. Consequently, although such a score may predict correctness empirically, it need not carry the information from which the posterior confidence could, in principle, be computed.

\begin{figure}[!t]
    \centering
    \includegraphics[width=0.8\textwidth]{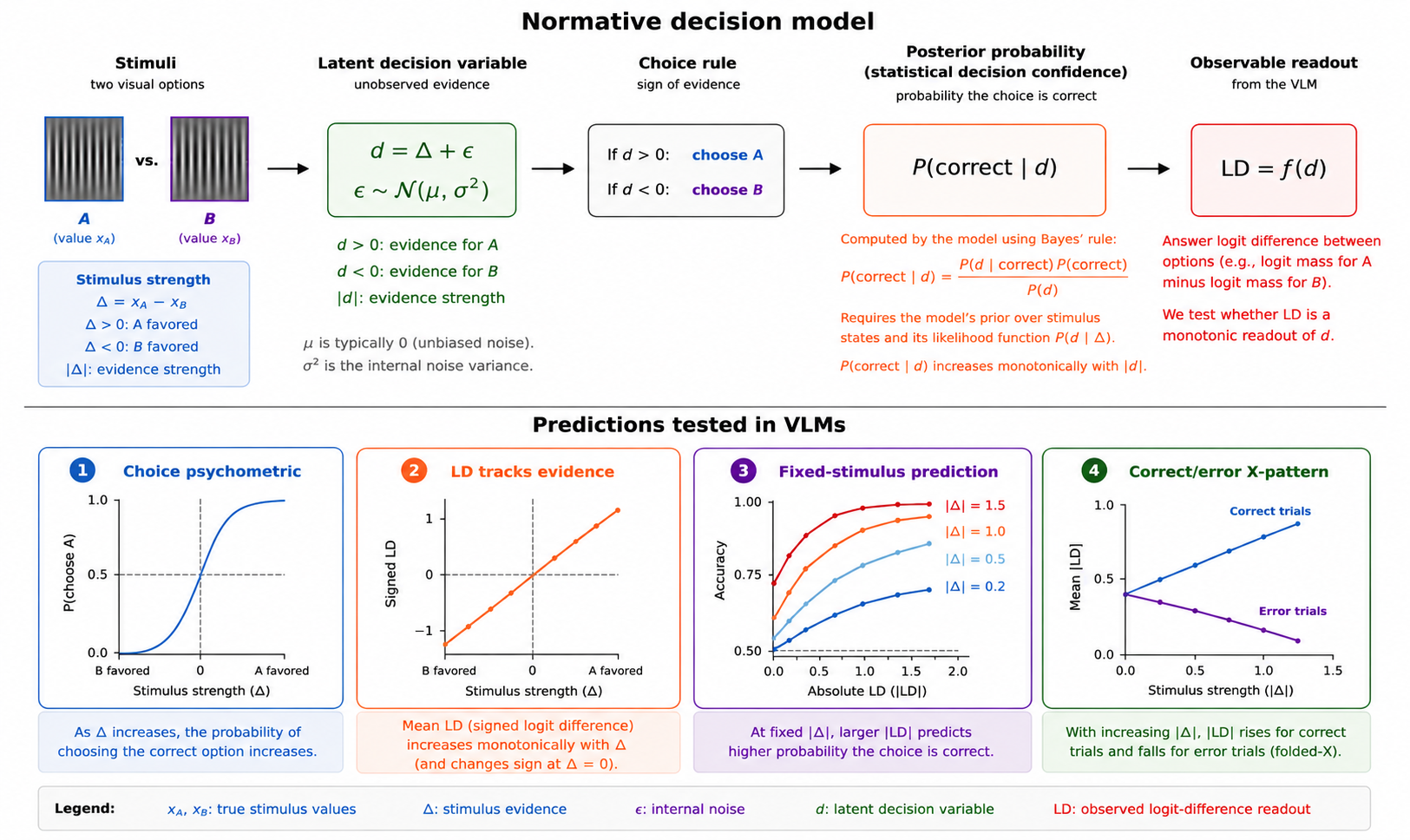}
\caption{\textbf{Schematic of the normative perceptual decision model used to define statistical decision confidence (SDC).}
The experimenter specifies a signed stimulus-evidence variable, \(\Delta\), which quantifies the evidence favouring one response over the other. In the normative observer, this evidence is encoded with internal noise to produce a latent decision variable, \(d=\Delta+\epsilon\), where \(\epsilon\) denotes internal decision noise. Consequently, a fixed stimulus strength induces a \emph{distribution} of latent decision values rather than a single value. The sign of \(d\) determines the choice, while its magnitude determines decision strength: larger \(|d|\) corresponds to stronger evidence for the chosen option and, under the normative model, is sufficient to determine the posterior probability that the current choice is correct, \(P(\mathrm{correct}\mid d)\). A deterministic multimodal VLM produces identical logits for identical inputs and therefore has no trial-to-trial encoding noise of this kind. Instead, images that share the same objective stimulus strength but differ in task-irrelevant (nuisance) features produce different internal evidence. This supplies, across stimulus exemplars, the variation in latent decision evidence that sensory noise supplies across repeated trials in the normative observer (main text). Because the latent decision variable is not directly observable, we ask whether the answer-logit difference (LD) behaves as a monotonic readout of it. The normative model predicts four qualitative signatures tested in the main text: (1) a psychometric function relating choice monotonically to signed stimulus strength; (2) signed LD varying monotonically with signed stimulus strength; (3) larger \(|LD|\) predicting a higher probability that the current choice is correct within fixed stimulus-strength bins; and (4) the folded-X pattern, in which confidence increases with stimulus strength on correct trials but decreases on error trials, because high-strength errors occur only when noise pushes the latent decision variable just beyond the decision boundary. This normative model is not intended as a unique mechanistic account; rather, it is one member of a broad class of noisy decision models—including drift-diffusion and sequential-sampling models \citep{bogacz2006physics}—in which a single latent evidence variable is sufficient in principle to determine both choice and the posterior confidence.}
\label{fig:SDC_predictions}
\end{figure}

We address this question through the framework of statistical decision confidence (SDC), developed in computational neuroscience \citep{kepecs2008neural, kepecs2012computational, hangya2016mathematical, sanders2016signatures}(see Methods for description of the ideal observer model and Figure~\ref{fig:SDC_predictions}). Under this framework, confidence is defined normatively as the posterior probability that the chosen option is correct given the latent decision variable---a noisy internal representation of the evidence in the stimulus favouring one alternative over the other. Critically, under the SDC framework this latent decision variable is a \emph{sufficient statistic} for the posterior: once it is known, the posterior probability of correctness is fully determined. Consequently, any signal that faithfully reads out this variable carries the information that is, in principle, sufficient to compute the posterior confidence---rather than merely indicating how strongly the model prefers one answer over another.

Our central question is therefore whether the observable answer-logit difference behaves like such a latent decision variable. If it does, answer logits are more than empirical predictors of correctness: they read out a quantity that is, in principle, sufficient to compute the posterior probability that the answer is correct. The statistical decision confidence framework provides a concrete way to test this hypothesis, making qualitative predictions about the relationship between the decision variable, confidence, and behaviour that can be evaluated empirically.

To investigate this hypothesis, we turn to simple perceptual decision tasks, where the experimenter explicitly specifies the task-relevant stimulus-evidence axis and thereby the normative process model linking evidence, choice, and confidence (Figure~\ref{fig:SDC_predictions})\citep{kiani2009representation, kepecs2008neural, lak2014orbitofrontal}. We use binary visual discrimination tasks inspired by classic experimental paradigms in animals \citep{kepecs2008neural, kiani2009representation, masset2020behavior}. On each trial, the stimulus favours one of two possible responses, and the strength of this evidence is varied systematically from ambiguous to easy, yielding a one-dimensional stimulus-strength axis analogous to classic perceptual decision experiments. This provides a normative benchmark against which the computational role of answer logits can be evaluated. Throughout the paper, we use the term \emph{stimulus strength} to refer to this experimenter-defined evidence favouring one response over the other.

The latent decision variable is not directly observable. Instead, we observe the logits assigned to the candidate answer tokens. We therefore use the answer-logit difference, $\mathrm{LD} = \ell_A - \ell_B$, as a candidate readout of the latent decision variable. The sign of LD indicates which answer the model favours (positive favours (A), negative favours (B), and values near zero indicate indifference), while its magnitude indicates how strongly the model favours the chosen answer over the alternative. The central question of this study is whether LD merely reflects a heuristic score that specifies the model's preference between competing responses, or whether it behaves as a readout of the latent decision variable---one that is sufficient in principle to compute the posterior confidence.

We test four signatures of statistical decision confidence, ordered from weakest to strongest, each illustrated by the normative model in Figure~\ref{fig:SDC_predictions} \citep{kepecs2008neural, hangya2016mathematical, sanders2016signatures}.

\emph{(1) Psychometric function.} As the stimulus increasingly favours answer (A), the probability of choosing (A) should increase monotonically, establishing that the model's choices are governed by the experimenter-defined stimulus evidence.

\emph{(2) Signed evidence.} Signed LD should vary monotonically with signed stimulus strength. This asks whether LD is a graded readout of the evidence favouring one answer over the other, with stimuli that increasingly favour one option producing progressively more positive LD and stimuli favouring the other producing progressively more negative LD. The relationship need not be linear. Together, these first two signatures concern the \emph{encoding} of evidence: they establish whether objective stimulus evidence is systematically represented in the model's decision variable. Importantly, these signatures are consistent both with a normative latent decision variable and with heuristic decision signals that preserve the ordering of stimulus evidence.

\emph{(3) Trial-specific confidence.} LD magnitude should predict correctness even when stimulus strength is held fixed. This is a key test: among trials of equal objective difficulty, choices made with larger $|\mathrm{LD}|$ should be more likely to be correct. If LD is a readout of the latent decision variable, statistical decision confidence predicts this positive within-strength relationship because trial-to-trial fluctuations in the decision variable determine the probability that the current choice is correct.

\emph{(4) Folded-X pattern.} As stimulus strength increases, confidence should increase on correct trials but \emph{decrease} on error trials, producing the characteristic folded-X pattern observed in perceptual decision studies \citep{kepecs2008neural,sanders2016signatures,hirokawa2019frontal}. Under statistical decision confidence, strong stimuli usually push the noisy decision variable far onto the correct side of the boundary, yielding high-confidence correct choices. However, when a strong stimulus nonetheless produces an error, the noisy decision variable must have crossed the decision boundary by only a small margin, yielding low-confidence errors.

Together, these final two signatures ask whether LD behaves as the latent decision variable prescribed by the normative model, rather than merely encoding stimulus evidence. They test whether trial-to-trial fluctuations in LD have the structure expected if LD is a readout of the latent decision variable---the variable that is sufficient in principle to compute the posterior probability that the current choice is correct. Such behaviour is not generally expected of heuristic decision signals that incorporate task-irrelevant information or combine task-relevant information in a manner that departs from the normative process model, since in these cases trial-to-trial variation in LD no longer reflects only the latent task-relevant evidence specified by the normative model. Collectively, the four signatures therefore progress from establishing that LD systematically encodes stimulus evidence to testing whether it reads out the latent decision variable prescribed by statistical decision confidence.

The final two signatures require trial-to-trial variation in the latent decision variable while objective stimulus strength is held constant. In biological observers, this variation arises naturally because repeated presentations of the same stimulus produce different internal sensory measurements, and therefore different values of the latent decision variable. Deterministic multimodal LLMs, however, do not exhibit this form of variability: identical inputs produce identical internal computations and identical answer logits. To recover the analogous regime, we generate multiple stimulus exemplars at each stimulus-strength level. Each exemplar preserves the same experimenter-controlled stimulus evidence favouring one response over the other while randomizing task-irrelevant (``nuisance'') properties of the image, such as the precise spatial arrangement of the visual elements. This parallels classic perceptual paradigms such as random-dot motion, where motion coherence fixes objective stimulus strength while the particular dot realization is resampled on every trial \citep{kiani2009representation}. Although these nuisance manipulations preserve objective stimulus strength, they induce trial-to-trial variation in its latent decision signal. Under the normative model, this variation plays the same role as sensory noise in the ideal observer, producing a distribution of latent decision values at fixed stimulus strength. Consequently, any relationship between $|\mathrm{LD}|$ and correctness within a stimulus-strength level reflects trial-specific variation in the model's internal decision signal rather than differences in objective difficulty, allowing signatures (3) and (4) to be tested in deterministic multimodal LLMs.

Pinning down the computational basis of confidence in multimodal LLMs is a fundamental step toward understanding what model confidence represents. Rather than asking simply whether confidence signals predict correctness, we ask whether they behave as readouts of the latent decision variable---one that is sufficient in principle to compute the posterior confidence. Simple perceptual decisions provide a uniquely tractable setting in which to answer this question because the evidence supporting each alternative is experimentally controlled and the predictions of statistical decision confidence are well defined. Establishing whether answer logits satisfy these predictions in this simplest setting provides a baseline against which confidence in reasoning, factual recall, and complex visual question answering can be interpreted.

\section*{Results}
We organize the results around the four signatures of statistical decision confidence introduced above and illustrated in Figure~\ref{fig:SDC_predictions} \citep{kepecs2008neural, hangya2016mathematical, sanders2016signatures}. The first two signatures concern the \emph{encoding} of evidence: whether choice and signed LD vary monotonically with the experimenter-defined stimulus-strength axis. The latter two test whether LD behaves as the latent decision variable prescribed by the normative model. Specifically, we ask whether, at fixed objective stimulus strength, larger $|\mathrm{LD}|$ predicts a higher probability that the current choice is correct, and whether correct and error trials exhibit the characteristic folded-X geometry. For the fixed-strength analysis, the primary quantitative result is the pooled logistic regression controlling for stimulus-strength bin; the binned plots are provided as visualizations of the same effect.

\subsection*{Qwen: Evidence Encoding and Latent Decision Variable Signatures in Perceptual Tasks}
\begin{figure}[!t]
    \centering
    \includegraphics[angle = -90, width=0.8\textwidth]{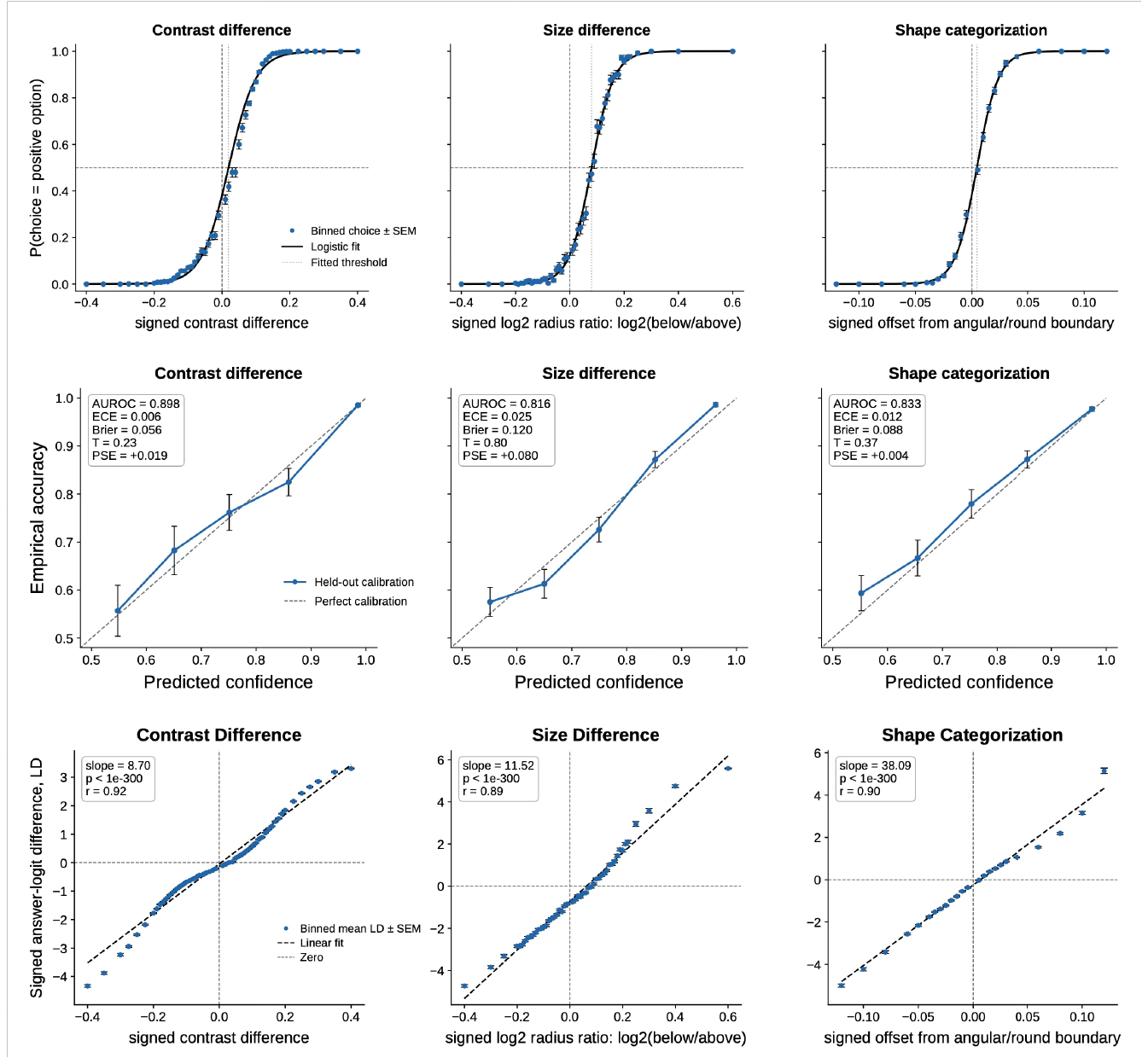}
\caption{\textbf{Qwen 2.5 7B: Choice, calibration, and evidence encoding across visual decision tasks.}
Top: Psychometric curves for the three perceptual tasks. Points show the empirical probability of choosing the positive-coded option, with SEM error bars; solid lines show fitted logistic psychometric curves. The signed stimulus axis is defined so that positive values favour the positive-coded response: higher contrast/right patch for contrast, lower circle for size, and round for shape categorization. In the size task, Qwen showed a substantial bias against the BELOW response: the fitted point of subjective equality was shifted to approximately $+0.080$ signed-size units, and at the objective boundary the model was predicted to choose BELOW on only $\approx 12.5\%$ of trials. Middle: Held-out calibration curves for answer-logit confidence, computed as $\sigma(|LD|/T)$, where $LD$ is the answer-logit difference and $T$ is fit on non-held-out trials by minimizing ECE. Calibration metrics are evaluated on 2{,}000 held-out trials per task; insets report AUROC, ECE, Brier score, and fitted temperature. Bottom: Strong linear correlation between signed LD and signed stimulus strength across tasks.}
\label{fig:psychom_calib}
\end{figure}

\begin{figure}[!t]
    \centering
    \includegraphics[angle = -90, width=0.8\textwidth]{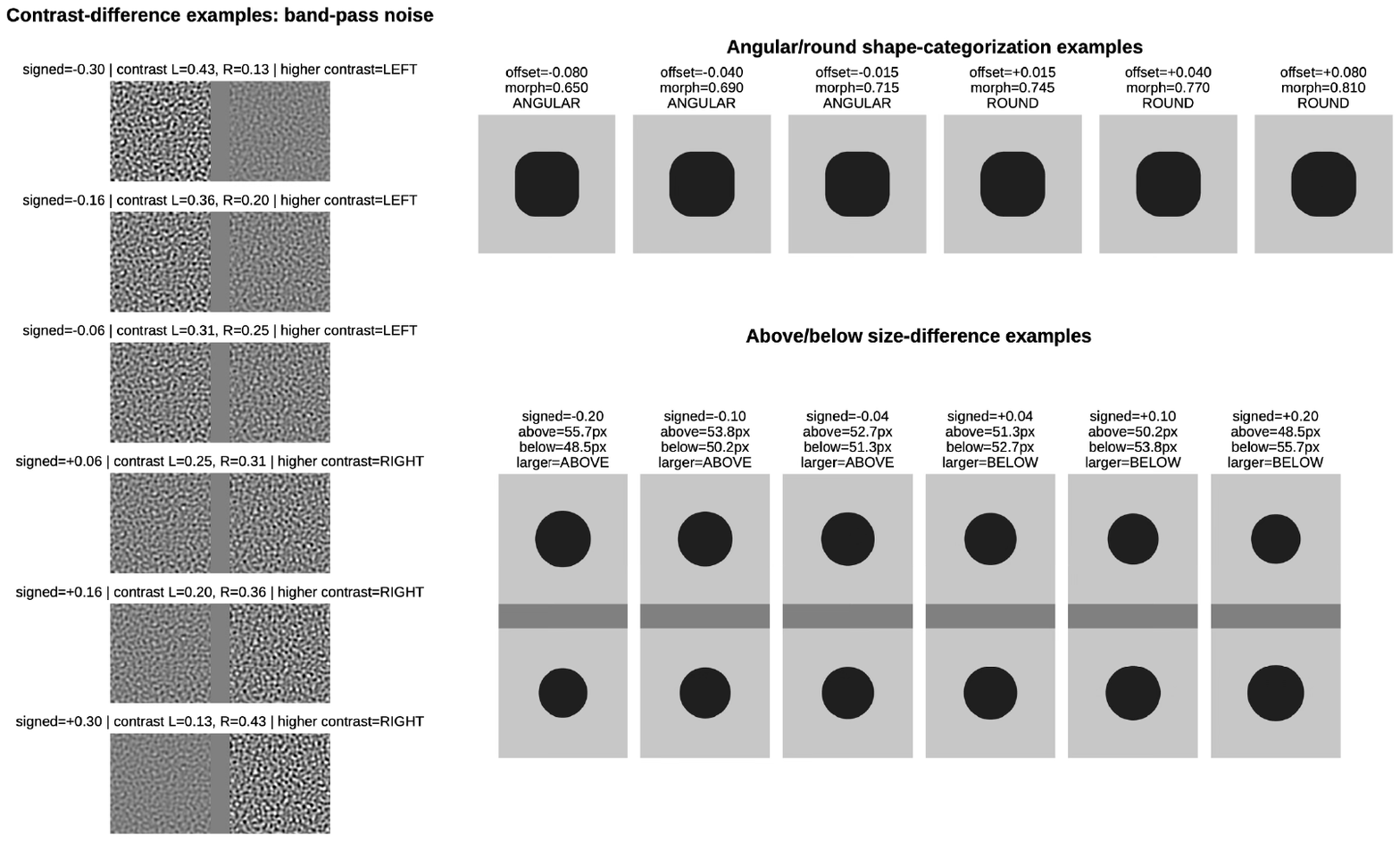}
\caption{\textbf{Example stimuli from simple perceptual decision making tasks.}In each task, the signed stimulus variable is defined so that positive values favour the positive-coded response, and the answer-logit difference, $LD$, is defined as the logit of the positive-coded response minus the logit of the negative-coded response. Thus $LD>0$ favours the positive-coded option, and $|\mathrm{signed}|$ indexes distance from the decision boundary. In the contrast task, two band-pass filtered noise patches were compared by contrast, with positive signed values indicating higher contrast on the right. In the size task, two vertically arranged circles were compared by radius, with signed value $\log_2(r_\mathrm{below}/r_\mathrm{above})$. In the shape task, a rounded-rectangle continuum was judged as angular versus round; signed offset was defined relative to the empirically estimated angular–round boundary.}
\label{fig:example_stimuli}
\end{figure}

\paragraph{Encoding of Evidence.}
Across the three synthetic perceptual decision tasks (see example stimuli in Figure~\ref{fig:example_stimuli}), the model demonstrated orderly psychometric behavior, satisfying the first core signature of evidence encoding. Choice probability increased monotonically with signed stimulus strength, and logistic fits effectively captured the behavioral transitions (Figure~\ref{fig:psychom_calib}). While objective performance was robust across the board, the model exhibited varying degrees of task-specific response bias—most notably a pronounced bias against the \texttt{BELOW} response in the size task—which are quantified by the shifts in the point of subjective equality (PSE) detailed in Table~\ref{tab:qwen_perceptual_summary}(A).

Crucially, the model also satisfied the second encoding signature: the signed answer-logit difference ($LD$) showed a strong linear correlation with signed stimulus strength. Mean signed $LD$ varied linearly with the physical evidence, crossing near the psychometric boundary and reversing sign across the two stimulus classes. This confirms that the model's answer logits carry a graded, signed representation of task-relevant evidence reflecting both the direction and strength of the sensory evidence. As a basic check, a single-temperature transformation of $|\mathrm{LD}|$ was both discriminative and well-calibrated on held-out trials across all three tasks (Figure~\ref{fig:psychom_calib}, middle)---demonstrating that LD is an accurate empirical predictor of correctness, but not yet establishing that it reads out the latent decision variable in the sense required by statistical decision confidence.

\paragraph{Latent decision variable signatures.}
This signed evidence code provides the necessary substrate to test the stricter readout predictions of statistical decision confidence in the subsequent analyses: specifically, whether $|LD|$ predicts correctness when objective stimulus difficulty is held constant, and whether it exhibits the signature folded-X correct/error interaction.

\paragraph{Answer-logit confidence predicts correctness at fixed stimulus strength.}
Even before controlling for difficulty, $|LD|$ predicted correctness beyond stimulus strength: across all three tasks it was the stronger predictor, and adding it to a stimulus-strength model significantly improved fit (contrast AUROC $0.864 \to 0.907$, combined $0.926$; all likelihood-ratio tests $p < 10^{-100}$). The stronger latent decision variable signature asks whether LD carries trial-specific decision evidence, rather than simply varying with objective stimulus strength. Because easier stimuli produce both larger $|LD|$ and higher accuracy, we asked whether $|LD|$ predicts correctness with stimulus strength held fixed, fitting logistic regressions of correctness on $|LD|$ with fixed effects for stimulus-strength bin, $\mathrm{correct} \sim C(|\mathrm{stimulus}|, \mathrm{bin}) + |LD|$. This removes the main effect of difficulty; the residual trial-to-trial variation arises from nuisance variation in the stimuli (independent noise realizations and positional jitter), which perturbs the model's internal evidence even at fixed nominal strength---the analogue of the internal noise exploited by fixed-stimulus confidence tests in animals \citep{kepecs2008neural}.

\begin{figure}[!t]
    \centering
    \includegraphics[angle = -90, width=1\textwidth]{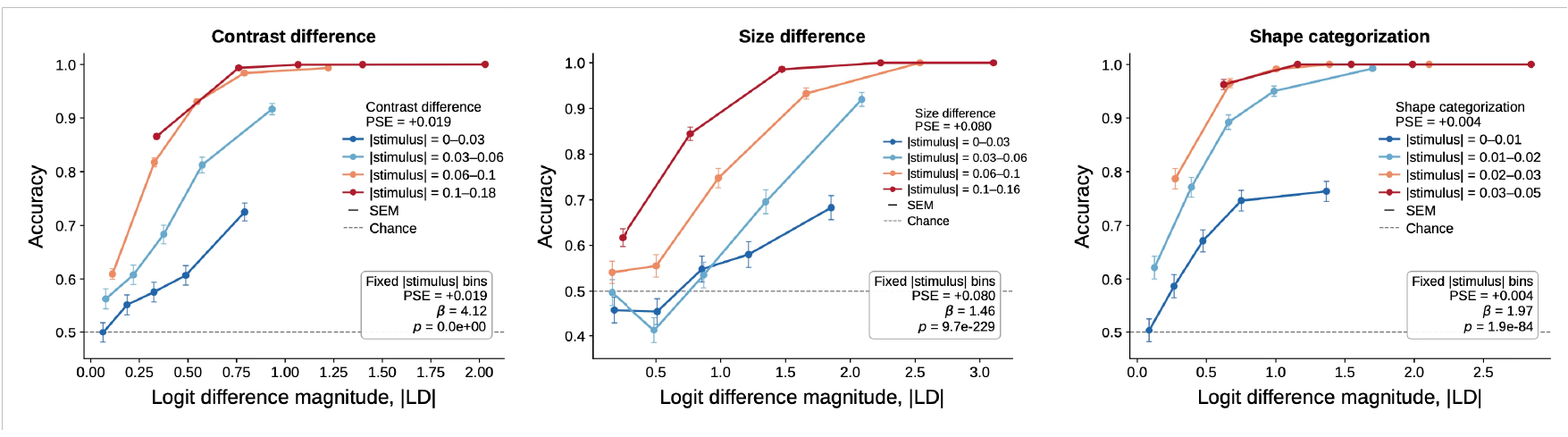}
\caption{\textbf{Qwen 2.5 7B: Answer-logit confidence predicts correctness within fixed stimulus-strength bins.}
Trials were grouped by absolute objective stimulus strength, $|\mathrm{signed}|$, and then binned by logit-difference magnitude within each strength bin. Points show empirical accuracy in each logit-difference bin, connected separately for each fixed stimulus-strength range. Insets report pooled logistic regressions predicting correctness from logit-difference magnitude while controlling for fixed stimulus-strength bin, $\mathrm{correct} \sim C(|\mathrm{stimulus}|, \mathrm{bin}) + |LD|$. Positive coefficients indicate that trial-to-trial variation in answer-logit confidence predicts correctness beyond physical stimulus difficulty. Error bars are SEM.}
\label{fig:fixed_ss}
\end{figure}

Across all three tasks, $|LD|$ remained a strong predictor of correctness within fixed stimulus-strength bins (Figure~\ref{fig:fixed_ss}; Table~\ref{tab:qwen_perceptual_summary}(B)). The effect was large and highly significant in every task, strongest for contrast ($\beta = 4.12$, $z = 53.42$, $p < 10^{-300}$): even with objective difficulty held fixed, trials with larger answer-logit differences were substantially more likely to be correct---establishing that Qwen’s LD carries trial-specific decision evidence beyond objective stimulus strength.

\paragraph{Correct and error trials show the folded-X.}
The strictest signature separates correct and error trials: statistical decision confidence predicts that, as stimulus strength increases, $|LD|$ rises on correct trials but falls on error trials, producing the folded-X (intuition and schematic in Figure~\ref{fig:SDC_predictions}). We tested this by regressing $|LD|$ on absolute stimulus strength, correctness, and their interaction, $|LD| \sim |\Delta| \times \mathrm{Correct}$.

\begin{figure}[!t]
    \centering
    \includegraphics[angle = -90, width=1\textwidth]{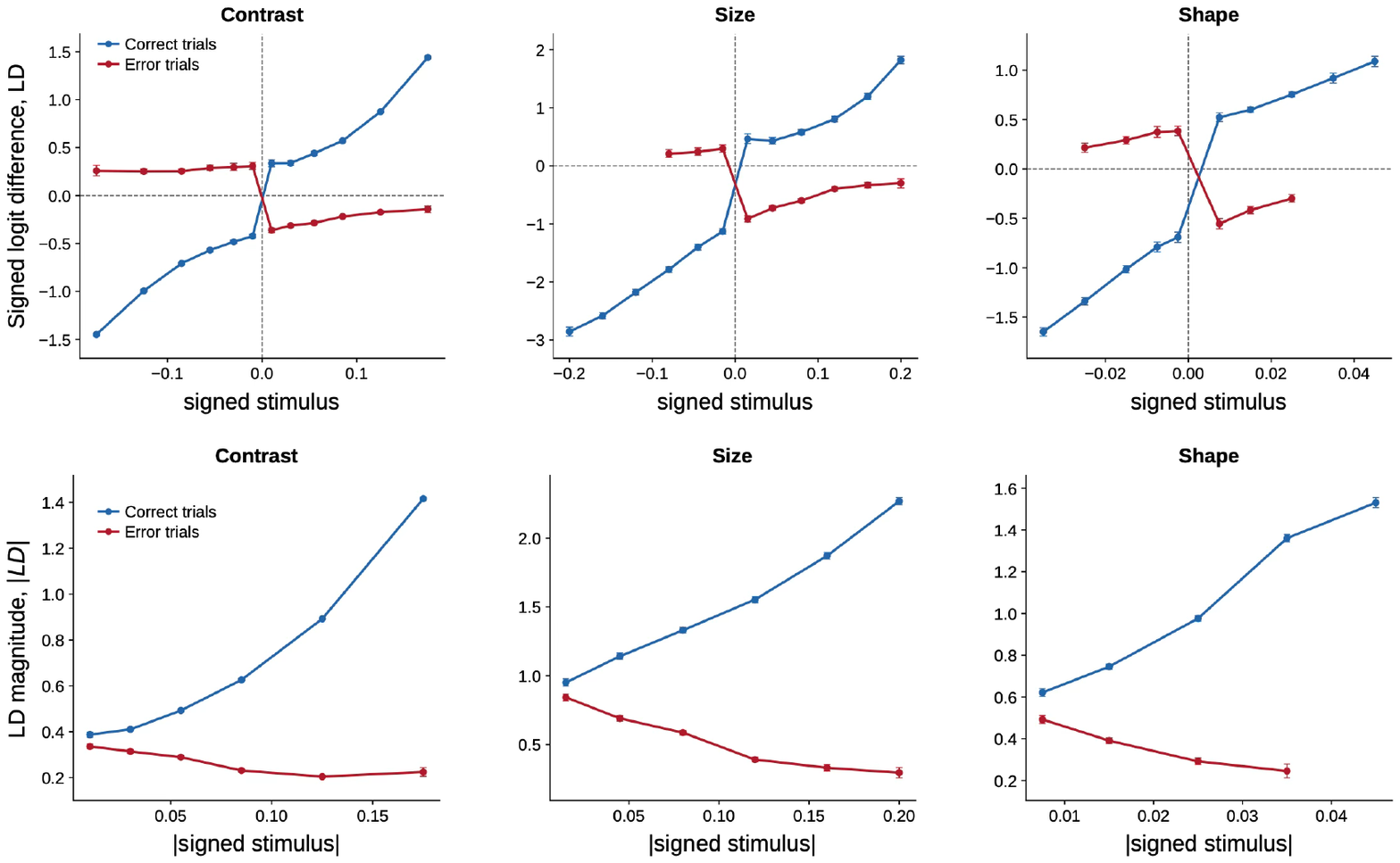}
\caption{\textbf{Qwen 2.5 7B: Decision-strength and signed evidence structure in answer logits.}
Top: Full signed X-pattern. The task-specific answer-logit difference \(LD\) is plotted against the signed stimulus variable, separately for correct and error trials. Correct trials lie on the stimulus-congruent side of \(LD=0\), whereas error trials lie on the opposite side. Middle: Calibrated confidence metrics using temperature scaling. Bottom: Logit-difference magnitude, \(|LD|\), plotted against absolute stimulus strength, \(|\mathrm{signed}|\). Correct trials show increasing \(|LD|\) with stimulus strength, whereas error trials remain lower and occur mainly near the decision boundary. Points show binned means with bootstrap 95\% confidence intervals.}
\label{fig:X_composite}
\end{figure}

Across all three tasks the interaction was strongly positive (Figure~\ref{fig:X_composite}; Table~\ref{tab:qwen_perceptual_summary}(C)): the slope of $|LD|$ on stimulus strength was positive for correct trials and negative for error trials (contrast, $+8.08$ vs.\ $-1.11$; the same pattern held for size and shape). These opposing slopes yield the folded-X. The folded-X is not a generic consequence of a signal predicting correctness or being well calibrated. Rather, it is the specific geometry predicted when confidence is a monotonic function of the latent decision variable. Across the perceptual tasks, Qwen satisfies all four signatures, indicating that answer-logit difference behaves as a readout of the latent decision variable---one that, under the experimenter-defined normative process model, is sufficient in principle to compute the posterior confidence. This provides the baseline against which we interpret the more complex tasks that follow.

\paragraph{Replication across models.}
The same signatures held in two further non-reasoning models, Gemma 3 12B (Figures~\ref{fig:gemma3_psychom},~\ref{fig:fixed_ss_gemma3},~\ref{fig:Gemma_X_composite}; Table~\ref{tab:gemma3_12b_summary}) and Gemma 4 12B (Figure~\ref{fig:Gemma4_composite}; Table~\ref{tab:gemma4_12b_contrast_shape_summary}), and in a reasoning model, Gemini Flash 3 (Figure~\ref{fig:Fierce_Falcon_composite}; Table~\ref{tab:fiercefalcon_contrast_90}; see Supplemental Results). In each, $|LD|$ predicted correctness within fixed stimulus-strength bins, and the folded-X was present in every model--task cell with a single exception; the Gemma models additionally reproduced the signed-LD evidence code. That exception was the Gemma 3 12B contrast error branch, which was flat rather than falling (Table~\ref{tab:gemma3_12b_summary}(D))---not, by itself, evidence against an SDC-like signal, as we take up in the Supplemental Results. For Gemma 4 12B the size task was excluded because objective performance was near chance, dominated by a response bias rather than graded uncertainty.

\paragraph{Summary.}
Across every model--task combination but one---three non-reasoning models and one reasoning model---answer-logit difference satisfied the discriminating signatures of statistical decision confidence: it predicted correctness at fixed stimulus strength and, in all but one case, exhibited the normative folded-X. The sole exception, Gemma~3 on the contrast task, departed only in the strictest signature, showing a flat rather than inverted error branch while still exhibiting strong within-strength prediction of correctness. Taken together, these results show that, in simple perceptual decisions, answer logits behave not merely as heuristic preference scores that predict correctness empirically, but as monotonic readouts of a latent decision variable. Consequently, they read out a quantity that, under the experimenter-defined normative process model, is sufficient in principle to compute the posterior confidence.

This conclusion concerns computational function rather than implementation. The normative model in Figure~\ref{fig:SDC_predictions} is not intended as a literal description of how multimodal LLMs compute their decisions. Rather, it specifies the functional properties of any signal that represents the posterior probability that the current choice is correct. Satisfying the four signatures therefore does not imply that models implement the particular computations of the normative observer. Instead, it establishes that the observable answer-logit difference behaves as though it were a readout of a latent decision variable with the same computational role: a representation from which the posterior probability that the current choice is correct can, in principle, be computed.

\paragraph{Abstention selectively identifies high-confidence decisions.}
We next asked whether the LD signal, which the preceding analyses identified as a readout of the latent decision variable, is behaviourally consequential---building on evidence that language models causally use confidence to drive behaviour \citep{kumaran2026causal}. After making the initial forced-choice contrast judgment, the model was given a second-stage decision: either commit its previous answer or abstain, and was instructed to abstain unless it believed its answer had at least a 60\% chance of being correct (see Methods). The resulting policy was conservative and highly selective. The model abstained on 87.7\% of trials, committing only its highest-confidence responses. Accuracy on these committed trials increased from the baseline forced-choice accuracy of 92.2\% to 99.95\%, while the original first-stage answers on abstained trials remained close to baseline accuracy (91.2\%; Table~\ref{tab:abstention_summary}). Thus, abstention selectively retained the most reliable decisions while rejecting those with low answer-logit confidence (low $|LD|$), which were disproportionately error-prone.

\begin{figure}[!t]
    \centering
    \includegraphics[angle = 0, width=0.5\textwidth]{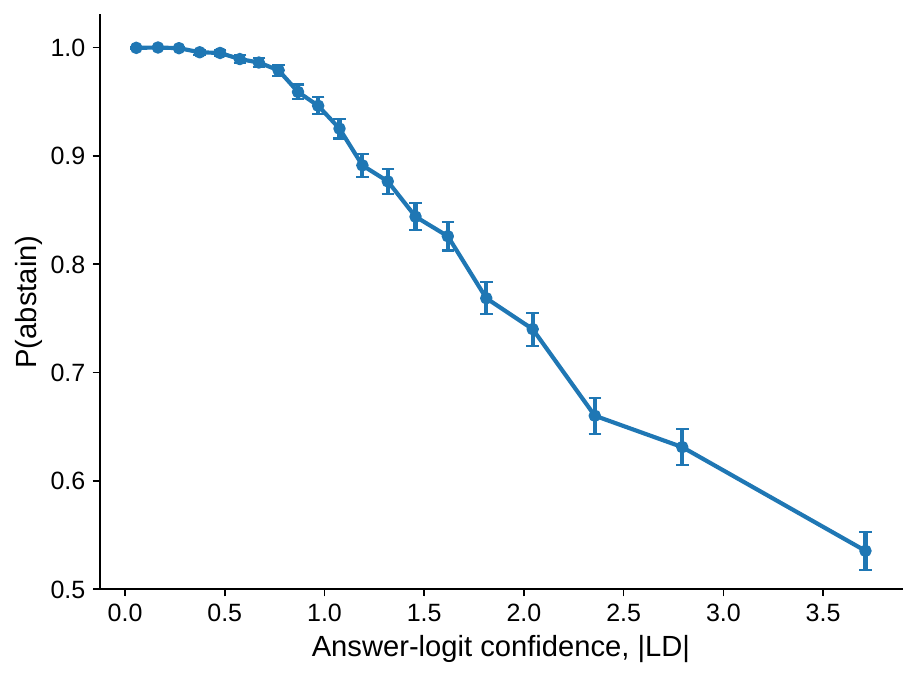}
\caption{Answer-logit confidence during the perceptual contrast task (LD) predicts abstention behavior in Qwen 2.5 7B}
\label{fig:abstention_LD}
\end{figure}

Critically, abstention tracked the internal confidence signal rather than external difficulty. Answer-logit confidence predicted abstention better than physical stimulus strength (AUROC $0.840$ vs.\ $0.796$), and adding stimulus strength barely improved this ($0.843$), so $|LD|$ carried most of the information the policy used (Figure~\ref{fig:abstention_LD}). The effect survived difficulty controls: within fixed stimulus-strength bins, lower $|LD|$ still strongly predicted abstention (odds ratio $=2.38$, $p<10^{-300}$). And the model's policy closely approximated an idealized rule that abstains on the lowest-$|LD|$ trials---the two overlapped on 91.6\% of trials, and the optimal threshold improved committed accuracy by only $0.05$ percentage points. Abstention is therefore governed by the same answer-logit signal characterized above, and acts on it almost exactly as a simple confidence threshold would.

\paragraph{Generalization to memory-based decisions: city-population comparisons.}
To test whether these signatures extend beyond perception, we turned to a memory-based decision---judging which of two cities has the larger population, which requires retrieving and comparing stored world knowledge rather than reading a perceptual stimulus (the perceptual-to-memory extension studied in humans by \citealt{sanders2016signatures}). We drew $5{,}000$ cities (population $>1$M) from the SimpleMaps database, giving $50{,}000$ binary comparisons. Here the experimenter's stimulus-strength axis---the ground-truth population ratio---is only a noisy proxy for the evidence the model brings to the choice, since recalled city size, retrieval strength, and semantic salience vary from trial to trial.

\begin{figure}[!t]
\centering
\includegraphics[angle=-90, width=1\textwidth]{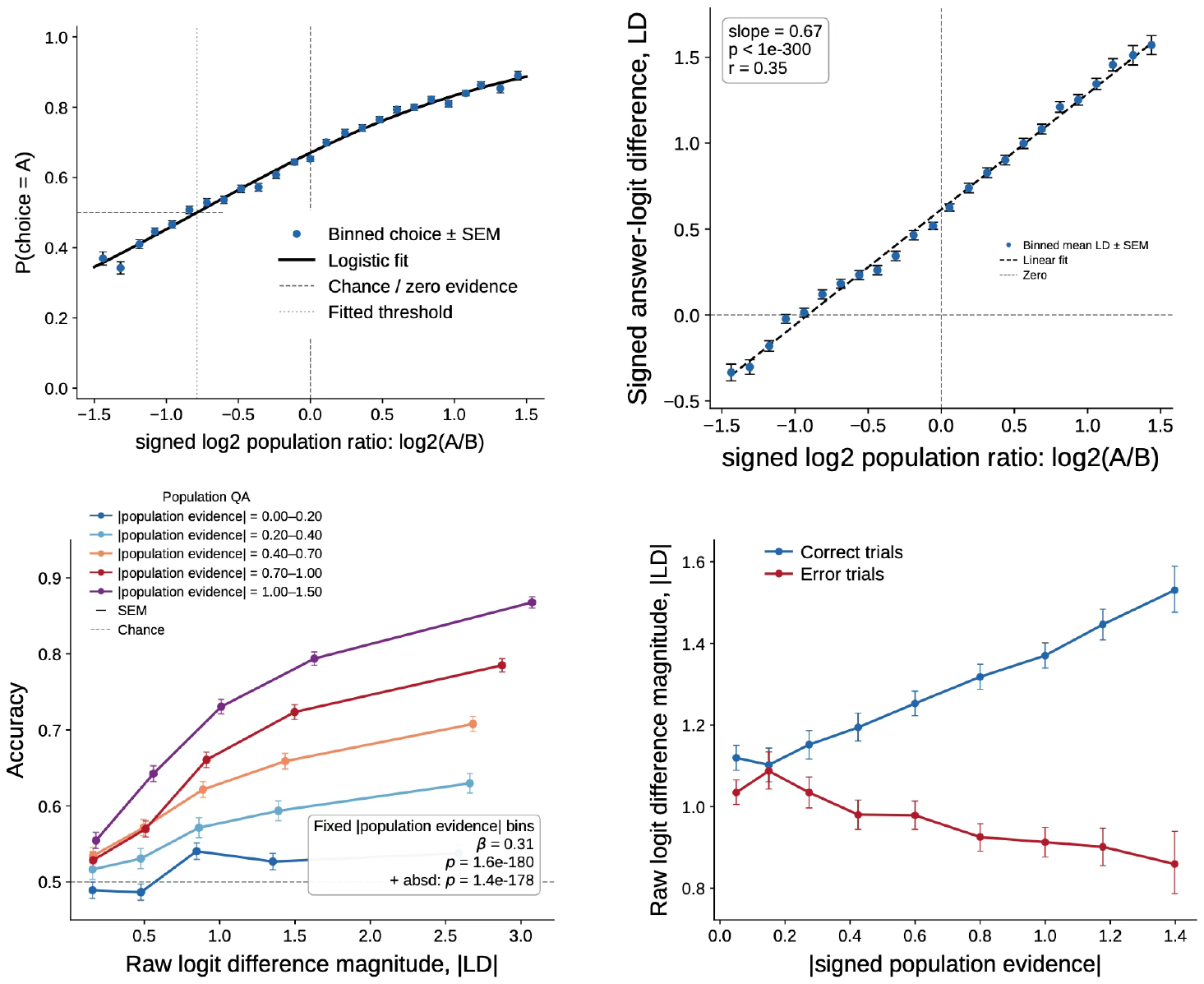}
\caption{\textbf{Qwen 2.5 7B, population QA: answer-logit confidence tracks correctness despite a shallow psychometric.}
Top left: psychometric curve; the signed stimulus is the $\log_2$ population ratio $\log_2(A/B)$. Choice was only weakly graded (slope $0.90$, pseudo-$R^2 = 0.069$) with a bias toward A (fitted threshold $-0.79$). Top right: signed $LD$ varied approximately linearly with binned population evidence---the mean encoding is intact despite the shallow psychometric. Bottom left: within fixed population-evidence bins, accuracy increased with $|LD|$. Bottom right: correct/error X-pattern; $|LD|$ rose with evidence on correct trials and fell on error trials. Error bars show SEM (psychometric, signed-LD, and fixed-bin panels) and bootstrap 95\% CIs (correct/error panel).}
\label{fig:Qwen_PopulationQA}
\end{figure}

This shows up in the encoding signatures. The psychometric function was shallow (accuracy $61.8\%$; slope $0.90$, pseudo-$R^2 = 0.069$), with a response bias toward option A (Table~\ref{tab:qwen_population_summary}(A)): objective population evidence was a weak \emph{trial-level} predictor of choice. But this reflects low signal-to-noise, not an absent mapping---signed $LD$ still varied approximately linearly with signed population evidence (Figure~\ref{fig:Qwen_PopulationQA}), so the second encoding signature, the mean mapping from objective evidence onto the decision variable, was intact. Encoding here is weak at the level of individual choices but faithful in the mean, the gap being large trial-to-trial noise in the recalled evidence.

Despite the shallow psychometric, both latent decision variable signatures held. Within fixed population-evidence bins, $|LD|$ predicted correctness (odds ratio $=1.36$, $p<10^{-180}$; Table~\ref{tab:qwen_population_summary}(B))---LD carried trial-specific decision evidence---and the full folded-X was present, with $|LD|$ rising with evidence on correct trials and falling on error trials (strongly positive interaction; Table~\ref{tab:qwen_population_summary}(C)). The same trial-to-trial noise that flattens the psychometric is what makes these latent decision variable signatures informative, since it is the within-strength fluctuation in evidence that they exploit.

The pattern replicated in Gemma 3 12B, where the psychometric was shallower still (pseudo-$R^2 = 0.049$) yet the folded-X was if anything sharper---a clearly negative error branch ($-0.589$) against a positive correct branch ($+0.678$; Figure~\ref{fig:gemma3_population}; Table~\ref{tab:gemma3_12b_population_summary}). This makes the memory task the clearest dissociation between objective stimulus evidence and the model's internal decision evidence. Here the experimenter-defined stimulus-strength axis predicts individual choices only weakly, yet LD continues to satisfy both latent decision variable signatures, indicating that its interpretation as a readout of the latent decision variable does not depend on a tightly coupled objective evidence axis.

\subsection*{CLEVR: Qwen 2.5 7B, Gemini Flash 3, and complex visual reasoning}
The perceptual tasks establish that answer logits can behave like a readout of the latent decision variable in settings where the normative process model can be well specified. We next ask whether this interpretation generalizes to complex visual reasoning, where the underlying decision process is no longer described by a simple perceptual evidence model.

\begin{figure}[!t]
\centering
\includegraphics[angle=-90, width=1\textwidth]{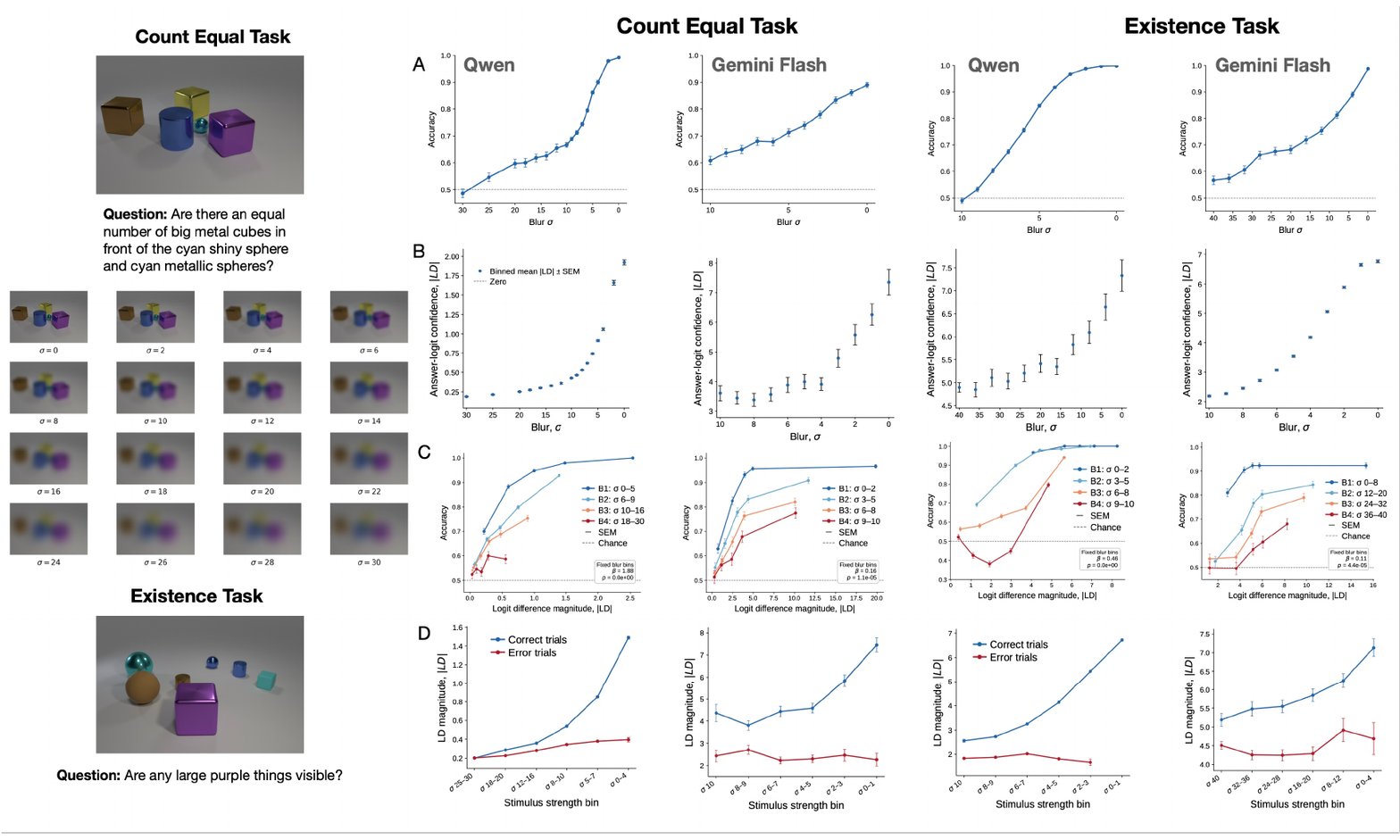}
\caption{\textbf{CLEVR count and existence tasks under parametric gaussian blur.}
Qwen 7B and Gemini Flash 3 were evaluated on CLEVR count-equality and object-existence questions while images were degraded with Gaussian blur. Larger blur $\sigma$ corresponds to lower stimulus strength; stimulus construction and blur levels are described in Methods. Example stimuli are shown on the left. For each task and model, row A shows accuracy as a function of blur, row B shows plot of $|\mathrm{LD}|$ against sigma (since sigma can only take on positive values), row C shows empirical accuracy as a function of $|\mathrm{LD}|$ within fixed blur bins, and row D shows the correct/error branch structure of $|\mathrm{LD}|$ across stimulus-strength bins. Accuracy-by-blur curves are binned for visual clarity; all statistical analyses use the individual trial/bin structure described in Methods. For Gemini Flash 3, analyses use the 90\% reasoning-cut readout. Across tasks and models, $|\mathrm{LD}|$ predicts correctness beyond blur, and correct-trial $|\mathrm{LD}|$ increases with stimulus strength. Correct branches have positive slopes throughout. Error-trial slopes are approximately flat in Gemini count and both existence analyses and Qwen existence, but positive in Qwen count. Full regression statistics are reported in Table~\ref{tab:clevr_xpattern_triallevel}.}
\label{fig:CLEVR_composite}
\end{figure}

We evaluated Qwen 2.5 7B and Gemini Flash 3 on CLEVR count-equality and object-existence questions while degrading images with parametric Gaussian blur (larger $\sigma$, lower stimulus strength; Figure~\ref{fig:CLEVR_composite}). One feature of this paradigm fundamentally changes what the SDC signatures can tell us. Unlike contrast or size discrimination, blur does not define a signed evidence axis: it removes information rather than favouring one answer over the other, and the mapping from blur to the model's latent decision variable is neither monotonic nor known. Consequently, the first two signatures—which rely on a well-defined evidence axis—do not transfer directly. What remains well defined is whether $|\mathrm{LD}|$ carries trial-specific information about whether the current choice is correct beyond objective image difficulty (signature~3), and how $|\mathrm{LD}|$ behaves on correct and error trials (signature~4).

The third latent decision variable signature remained robust. Adding $|\mathrm{LD}|$ to blur strength in nested logistic models improved discrimination of correct from error trials in every dataset, with AUROC gains of $+0.029$ to $+0.072$, confirmed by likelihood-ratio tests throughout (Table~\ref{tab:clevr_nested_fixed_bin_summary}), and the same conclusion was reached non-parametrically within fixed blur bins (Figure~\ref{fig:CLEVR_composite}). We next asked whether this effect could be explained by other measurable properties of the input. Adding scene and question controls—including the number of unique objects and colours, and spatial crowding (Methods)—improved prediction, particularly for counting, yet $|\mathrm{LD}|$ continued to add significant information beyond blur and these controls in all estimable nested models. In the fixed-blur analyses with the full control set, the $|\mathrm{LD}|$ effect remained significant for Qwen count, Qwen existence, and Gemini Flash 3 existence; the Gemini Flash 3 count model was not estimable owing to collinearity among the control variables, so we make no claim there. Thus, across all estimable models, answer-logit difference continued to carry information about whether the current choice was correct beyond the observable properties of the input.

The correct/error structure was more nuanced. On correct trials, $|\mathrm{LD}|$ increased with stimulus strength in all four analyses. On error trials, however, the slope was approximately flat (Gemini count, Qwen existence, Gemini existence) or modestly positive (Qwen count), never negative, although the correct$\times$error interaction remained significant throughout (Table~\ref{tab:clevr_xpattern_triallevel}; Figure~\ref{fig:CLEVR_composite}, row~D). Thus, $|\mathrm{LD}|$ continued to behave differently on correct and error trials, but did not exhibit the canonical folded-X.

Whether this constitutes a failure of statistical decision confidence cannot be determined from these data alone. The folded-X is a prediction of a specific normative process model in which the experimenter specifies both the task-relevant evidence axis and its mapping to the latent decision variable. In CLEVR, neither is known. Consequently, the normative posterior cannot be derived, and the folded-X is no longer a diagnostic prediction. We therefore cannot conclude that answer logits in CLEVR either satisfy or violate statistical decision confidence. The secure conclusion is the weaker one: answer-logit difference continues to carry trial-specific information about whether the current choice is correct beyond objective properties of the input, but whether it behaves as a normative latent decision variable remains unresolved until an appropriate process model is established (see Discussion).


\section*{Discussion}
We asked whether answer logits in multimodal LLMs merely provide useful predictors of correctness, or whether they warrant the stronger interpretation as readouts of the latent decision variable prescribed by statistical decision confidence (SDC). We addressed this question in three multimodal non-reasoning models (Qwen2.5-VL, Gemma~3, Gemma~4) and one multimodal reasoning model (Gemini Flash~3) across three decision regimes: low-level perceptual discrimination (contrast, size, shape), a memory-based decision (city-population comparison), and complex visual reasoning (CLEVR count and existence tasks under blur). Across the perceptual and memory tasks, and consistently across models, answer-logit difference (LD) satisfied all four qualitative signatures of SDC. LD encoded the signed stimulus evidence, predicted correctness beyond objective stimulus strength, and—with the single exception of Gemma~3 on the contrast task—exhibited the characteristic folded-X geometry predicted when confidence is a monotonic function of the latent decision variable. The same signal was also behaviourally consequential: in a second-stage abstention task, LD predicted the model's decision to commit or abstain better than objective stimulus strength, and the resulting policy closely approximated a simple threshold on LD. Together, these findings show that, in simple perceptual and memory-based settings, answer logits are more than heuristic preference scores that predict correctness empirically: they behave as monotonic readouts of the latent decision variable---a quantity that, under the experimenter-defined normative process model, is sufficient in principle to compute the posterior confidence.

A central issue is that statistical decision confidence is defined relative to a specified normative process model. In simple perceptual tasks, the experimenter defines both the task-relevant evidence variable and the process by which noisy evidence generates a latent decision variable. In more complex tasks, neither is generally known. Consequently, an apparent failure of the SDC signatures does not, by itself, imply that answer logits are non-normative. We believe the CLEVR results should be interpreted in this light. Crucially, answer-logit difference continued to carry trial-specific information about whether the model's current choice was correct beyond objective properties of the input, even after controlling for blur and measurable scene properties. What was absent was the folded-X. We do not interpret this as evidence against statistical decision confidence. Rather, the folded-X is a prediction of a specific normative process model in which the experimenter specifies both the task-relevant evidence axis and its mapping onto a latent decision variable. In CLEVR, neither is known. Consequently, the absence of the folded-X does not by itself distinguish between a failure of statistical decision confidence and a mismatch between the experimenter-defined process model and the computation actually performed by the model.

More generally, apparent departures from the experimenter-defined normative model may arise in at least three ways. First, the model may incorporate task-irrelevant information into its decision variable. A simple thought experiment illustrates how this can arise. Consider a recognition-memory task in which the model must decide which of two items has been encountered more often. The experimenter defines the relevant evidence as familiarity acquired during the experiment. Suppose, however, that the model cannot distinguish this from pre-existing real-world familiarity, so that its latent decision variable combines both sources of evidence (see Supplemental Discussion for details). The model's effective evidence variable therefore differs from the experimenter-defined evidence axis, and behaviour analysed with respect to experimental familiarity need no longer exhibit the signatures predicted by the normative model, even if confidence is computed normatively with respect to the model's own effective evidence variable. An analogous mismatch may underlie the CLEVR results if the model's effective evidence depends upon latent variables not captured by the experimenter-defined blur axis.

Second, the model may ignore task-relevant information specified by the experimenter. Third, it may combine the available evidence in a non-Bayesian (heuristic) fashion rather than computing the normative posterior. Any of these possibilities could account for the absence of the folded-X in CLEVR. Distinguishing between them requires developing explicit computational process models for complex reasoning tasks, analogous to those available for simple perceptual decisions.

It is also worth distinguishing the aim of the present framework from the focus of confidence calibration \citep{guo2017calibration, steyvers2025metacognition, xiong2023can, tian2023just, kadavath2022language}. Calibration seeks a mapping from a model's raw confidence signal to empirical correctness, so that reported confidence better matches the observed frequency of correct responses. This is valuable for prediction and deployment, but it does not by itself establish what the underlying signal represents. A well-calibrated confidence score may arise from a latent decision variable supporting the normative posterior, but it may equally arise from a heuristic preference signal that predicts correctness empirically. Statistical decision confidence asks a different question: whether the underlying signal itself behaves as a monotonic readout of the latent decision variable prescribed by a normative process model. Calibration therefore concerns the mapping between a confidence score and empirical accuracy, whereas the SDC framework concerns the computational interpretation of the confidence signal itself.

One qualification concerns the level of inference these results support, and it connects directly to a long-standing debate about human confidence: whether confidence reflects the posterior probability of being correct (the Bayesian confidence hypothesis, BCH; \citealp{pouget2016confidence, kepecs2012computational, adler2018limitations}) or the strength of the raw evidence itself, without an explicit probability computation (confidence in raw evidence space, CRES; \citealp{xue2024challenging}). Because our tasks use boundary-defined, non-overlapping response categories, they fall in the regime where the folded-X is a robust rather than fragile prediction of statistical decision confidence \citep{adler2018limitations}. Within this regime the four signatures establish that answer-logit difference behaves as a readout of the latent decision variable, a quantity sufficient in principle to compute the posterior. What they do not fix is whether $|LD|$ is that posterior or a monotone transform of it that does not renormalise for evidence reliability. Because the normative posterior is itself a monotonic (sigmoidal) function of the decision variable, any faithful readout of that variable---whether the calibrated posterior or the raw evidence---imposes the same ordering on trials and so reproduces all four signatures at any fixed level of reliability; what the signatures cannot see is whether the slope of that mapping tracks reliability. BCH and CRES therefore diverge only when reliability is manipulated independently of stimulus strength---under BCH the mapping from $|LD|$ to accuracy is reliability-invariant, because a calibrated posterior already absorbs reliability, whereas under CRES a fixed $|LD|$ corresponds to progressively lower accuracy as reliability falls, because raw evidence is not renormalised. Our single signed-strength axis varies evidence and reliability together and therefore cannot separate the two. Adjudicating them requires independently manipulating stimulus strength and evidence reliability, together with quantitative model comparison rather than qualitative signatures alone \citep{adler2018limitations, xue2024challenging}---an important direction that the framework developed here newly makes tractable for artificial systems.

Our results also bear on the distinction between first- and second-order theories of confidence \citep{pouget2016confidence, fleming2017self, mamassian2022modeling}. The present work concerns the confidence carried by answer logits themselves. We find that, in simple perceptual and memory-based decisions, answer-logit difference behaves as a readout of the latent decision variable prescribed by statistical decision confidence. This does not preclude the existence of additional confidence computations. Recent work suggests that verbally reported confidence and answer-logit confidence are dissociable signals that independently influence behaviour \citep{kumaran2026llms, kumaran2026causal, kumaran2026commitment}, implying that models may construct a second-stage evaluation of their own answers beyond the decision signal studied here. Rather than competing explanations, these signals may occupy different levels of the confidence hierarchy: answer-logit difference provides the decision-linked signal supporting statistical decision confidence, whereas verbal confidence may incorporate additional computations reflecting broader uncertainty about the answer. Understanding how these signals interact, when they diverge, and which ultimately governs behaviour remains an important direction for future work.

More broadly, our results suggest that statistical decision confidence provides a principled framework for understanding what model confidence represents, rather than simply how well it predicts correctness. Although we have focused primarily on visual perceptual decisions, the framework is not inherently visual. We further showed that the same qualitative signatures extend to a memory-based language task, and answer-logit differences are available whenever a model commits to a discrete response. The framework therefore provides a general approach for studying confidence in language models, multimodal agents, and more complex reasoning systems, provided an appropriate normative process model can be specified. Finally, the signatures studied here are behavioural. Identifying the internal representations that give rise to them would provide a mechanistic account of confidence analogous to the progression in neuroscience from behavioural signatures of confidence to neural representations of the underlying decision variable.

\clearpage
\section*{Methods}
\subsection*{Normative model}

The four signatures we test derive from a standard normative (ideal-observer) model of two-alternative decision making, schematised in Figure~\ref{fig:SDC_predictions}. The experimenter specifies a signed stimulus-evidence variable, $\Delta$, which quantifies the evidence favouring one alternative over the other. The observer does not access $\Delta$ directly, but instead forms a noisy internal representation,

\[
d=\Delta+\epsilon,
\]

where $\epsilon$ is internal decision noise. The latent decision variable $d$ therefore fluctuates from trial to trial even when the objective stimulus evidence is held fixed. These trial-to-trial fluctuations are essential: they make a fixed stimulus strength map to a distribution of latent decision values rather than a single value, and it is this variation that the latent decision variable signatures (3 and 4) exploit.

A deterministic VLM produces identical logits for identical inputs, so such variability does not arise from repeated presentation of the same image. Instead, it is supplied across stimulus exemplars: images with the same objective stimulus strength but differing in task-irrelevant (nuisance) features produce different internal evidence. The nuisance manipulations used for each task are described below.

The sign of $d$ determines the observer's choice. Confidence is defined by statistical decision confidence (SDC) as the posterior probability that this choice is correct,

\[
\mathrm{SDC}
=
P(\mathrm{correct}\mid d).
\]

Computing this posterior requires combining the prior distribution over stimulus states with the likelihood of observing the latent decision value under each possible stimulus state,

\[
P(\mathrm{correct}\mid d)
=
\frac{P(d\mid \mathrm{correct})P(\mathrm{correct})}
{P(d)},
\]

where the denominator marginalises over all possible stimulus states. Thus, the latent decision variable is a sufficient statistic for confidence: once $d$ is known, the posterior probability that the current choice is correct is fully determined. The posterior increases monotonically with $|d|$, so a single latent variable determines both the observer's choice (its sign) and its normative confidence (its magnitude).

Neither $d$ nor the observer's implied likelihood function are directly observable. Rather than attempting to recover the posterior itself, we ask whether the observable answer-logit difference behaves as a monotonic readout of the latent decision variable,

\[
\mathrm{LD}=f(d),
\]

where $f$ is monotonic. If so, LD inherits the qualitative signatures predicted for the latent decision variable---the psychometric function, signed-evidence encoding, fixed-strength correctness prediction, and the folded-X pattern---without requiring explicit recovery of the posterior.

\subsection*{Tasks}
\paragraph{Simple perceptual decision tasks}
We constructed three simple perceptual tasks that allowed precise experimental control over stimulus strength while providing large numbers of trials near perceptual decision boundaries. All tasks were evaluated using the same multimodal LLM and analysis pipeline. Each task generated a signed stimulus-strength variable, $\Delta$, where positive and negative values favoured opposite response categories. Model choices were analysed using response-token logits, allowing extraction of both categorical decisions and continuous decision variables.

\paragraph{Contrast discrimination.}
Stimuli consisted of two independent band-pass filtered noise patches presented side-by-side on a uniform grey background. Each patch was $256 \times 256$ pixels and the two patches were separated by a 24-pixel gap. Noise images were generated by sampling Gaussian white noise and applying a log-Gaussian band-pass filter in the Fourier domain. The filter was centred at 4 cycles/degree with a bandwidth of 1 octave. The filtered image was transformed back to the spatial domain, normalised to zero mean and unit variance, and scaled to the desired root-mean-square (RMS) contrast.

The baseline RMS contrast was fixed at $c_0 = 0.28$. On each trial, left and right contrasts were determined by a signed contrast difference parameter $\Delta$,
\[
c_L = c_0 - \frac{\Delta}{2},
\qquad
c_R = c_0 + \frac{\Delta}{2}.
\]
Positive values of $\Delta$ therefore indicated that the right patch had higher contrast, whereas negative values indicated that the left patch had higher contrast.

Stimulus strengths ranged approximately from -0.40 to +0.40. Sampling density was highest near $\Delta = 0$, with additional stimulus levels extending into easier discrimination regimes. Independent random seeds were used for the left and right noise patches on every trial.

The model was asked:
\begin{quote}
\emph{``This image shows two noise patches, LEFT and RIGHT. Which has HIGHER CONTRAST?''}
\end{quote}

Responses were given using two response letters. To eliminate confounding between perceptual side preferences and token preferences, the mapping between response letters and spatial locations was counterbalanced across trials. Half of trials used $M$ for left and $V$ for right, whereas the remaining trials used the reverse mapping.

Let $\ell_R$ and $\ell_L$ denote the total logit mass assigned to the tokens currently mapped to the right and left alternatives. The signed decision variable was
\[
d = \ell_R - \ell_L.
\]

\paragraph{Object-size discrimination.}
Stimuli consisted of two filled circles arranged vertically. Each circle was rendered within a $256 \times 256$ pixel image region, with the upper and lower regions separated by a 48-pixel gap. Circle luminance was fixed at 0.12 on a background luminance of 0.78. Shapes were rendered using $4\times$ supersampling followed by Lanczos downsampling to produce anti-aliased edges.

The baseline circle radius was $r_0 = 52$ pixels. Stimulus strength was parameterised using a signed log-radius ratio,
\[
\Delta = \log_2 \left( \frac{r_{\mathrm{below}}}{r_{\mathrm{above}}} \right).
\]
Circle radii were generated according to
\[
r_{\mathrm{above}} = r_0 2^{-\Delta/2},
\qquad
r_{\mathrm{below}} = r_0 2^{+\Delta/2}.
\]
This parameterisation preserved the geometric mean radius across trials while manipulating relative size. Positive values of $\Delta$ indicated that the lower circle was larger.

To prevent exploitation of trivial spatial cues, each circle received an independent random positional jitter sampled uniformly from $\pm 6$ pixels in both horizontal and vertical directions. Stimulus strengths ranged approximately from -0.40 to +0.60 log-radius units, with oversampling concentrated near the psychometric boundary.

The model was instructed:
\begin{quote}
\emph{``Two circles are shown vertically, one higher and one lower. Which circle is larger? Answer with exactly one word: ABOVE or BELOW.''}
\end{quote}

The signed decision variable was defined as
\[
d = \ell_{\mathrm{below}} - \ell_{\mathrm{above}},
\]
where $\ell_{\mathrm{below}}$ and $\ell_{\mathrm{above}}$ denote the logit masses assigned to the corresponding response tokens.

\paragraph{Shape categorization.}
Stimuli consisted of a single centrally presented rounded rectangle rendered within a $256 \times 256$ pixel image. Shapes were displayed using a background luminance of 0.78 and object luminance of 0.12. Anti-aliasing was implemented using $4\times$ supersampling followed by Lanczos downsampling.

The base side length was 120 pixels. Shape category was manipulated using a continuous morph parameter $m \in [0,1]$, where $m=0$ corresponded to a square-like angular shape and $m=1$ corresponded to a circle. The morph continuum was implemented by varying the corner radius of a rounded rectangle according to
\[
r_{\mathrm{corner}} = m \frac{s}{2},
\]
where $s$ denotes side length.

Stimulus strength was defined relative to each model's empirically estimated angular--round boundary $m^\ast$, as $\Delta = m - m^\ast$, so that $\Delta = 0$ marks the model's point of subjective equality ($m^\ast = 0.73$ for Qwen, $0.38$ for Gemma~3).
Negative values of $\Delta$ corresponded to shapes categorised as angular, whereas positive values corresponded to shapes categorised as round.

To introduce nuisance variation, side length was independently jittered by up to $\pm 6\%$ on each trial and stimulus position was jittered by up to $\pm 6$ pixels. Stimuli were sampled densely in the vicinity of the empirical category boundary, with $\Delta$ values extending approximately from -0.12 to +0.12.

The model was instructed:
\begin{quote}
\emph{``A single shape is shown. Is the shape more ANGULAR or more ROUND? Answer with exactly one word: ANGULAR or ROUND.''}
\end{quote}

The signed decision variable was
\[
d = \ell_{\mathrm{round}} - \ell_{\mathrm{angular}}.
\]

\paragraph{Abstention task: Qwen 2.5 7B}
To test whether decision-confidence signals predicted subsequent behavior, we conducted a second-stage abstention experiment using the contrast-discrimination task. Each abstention trial began with a previously generated forced-choice response from the model. The original image, question, and model answer were then presented back to the model in a multi-turn chat format. The model was asked to decide whether to output its previous answer to the user (\texttt{A}) or abstain (\texttt{B}).

The abstention prompt was:
\begin{quote}
\emph{``You now need to decide whether to output your previous answer to the user (A) or abstain (B).\\
If you don't think your answer has 60\% chance of being correct, choose B although this will disappoint the user.\\
Respond with exactly one letter.\\
A = output answer\\
B = abstain\\
Decision:''}
\end{quote}

The 60\% threshold was selected empirically. Preliminary experiments revealed that more permissive prompts resulted in near-universal commitment, whereas stricter prompts resulted in near-universal abstention. Our goal was not to optimize abstention performance but to place the model in a regime where both commitment and abstention occurred frequently enough to permit analysis of the relationship between internal confidence signals and overt behavior. The chosen threshold yielded substantial variability in abstention decisions, allowing us to test whether answer-logit confidence predicted the model's willingness to stand behind its previous answer. The generated abstention decision (\texttt{A} or \texttt{B}) was used for all behavioural analyses.

\paragraph{City-population binary-choice task.}
To test whether answer-logit confidence generalized beyond perceptual discrimination, we constructed a factual memory-based decision task using the SimpleMaps World Cities Database. The database provides city names, country information, geographic coordinates, and population estimates for cities and towns worldwide. We restricted the stimulus set to cities with population greater than one million people. This yielded $5,000$ unique cities.

Trials consisted of binary population comparisons between two cities. On each trial, the model was asked which of two cities had the larger population. The signed stimulus-strength variable was defined as
\[
\Delta = \log_2\left(\frac{\mathrm{population}_A}{\mathrm{population}_B}\right),
\]
so that $\Delta>0$ indicated that option A was correct and $\Delta<0$ indicated that option B was correct. The absolute stimulus strength, $|\Delta|$, therefore measured the population-ratio difference between the two cities. City pairs were sampled from the eligible city set to generate 50,000 binary-choice trials. Cities were allowed to appear in multiple different pairs, but each trial contained two distinct cities.

The prompt was:
\begin{quote}
\emph{``Which city has the higher population?\\
A. [city A]\\
B. [city B]\\
Answer with exactly one letter: A or B.''}
\end{quote}

The model's first-token logits were extracted at the answer position. Let $\ell_A$ and $\ell_B$ denote the logit masses assigned to the answer tokens \texttt{A} and \texttt{B}. We defined the signed decision variable as
\[
LD = \ell_A - \ell_B.
\]
Thus, positive values indicated evidence favouring option A and negative values indicated evidence favouring option B. Accuracy was computed by comparing the sign of $LD$ with the sign of $\Delta$, and answer-logit confidence was defined as $|LD|$.

\paragraph{CLEVR count and existence tasks}
CLEVR \citep{johnson2017clevr} is a synthetic visual-question-answering benchmark of rendered 3D scenes whose objects vary in shape, colour, material, and size, paired with compositional questions and ground-truth scene graphs and functional programs---a controlled, fully annotated structure that lets us select well-defined count and existence questions and compute objective scene-difficulty measures while degrading only the rendered image. We drew questions from the v1.0 training and validation splits. Scene annotations and programs were used only to select questions and, for the count task, to compute the two queried object counts; the model was always evaluated on the rendered image and natural-language question, never on the scene graph or program.

Both tasks used a two-letter forced-choice format, with the mapping between letters and response categories counterbalanced within each image--question pair (each letter assigned to each category once across the two presentations). We defined the semantic logit difference as the log-probability of the letter denoting the target category (\emph{equal} for count, \emph{yes} for existence) minus that of the alternative, averaged across the two letter mappings to remove fixed letter-identity biases. Questions were sampled deterministically (seed 42), and each image was evaluated across a range of Gaussian blur levels parameterised by filter radius $\sigma$.

For Qwen2.5-VL, which showed a strong affirmative-response bias that counterbalancing does not remove, analyses were restricted to yes-truth trials (equal counts in the count task; queried object present in existence), avoiding the regime in which responding was bias-dominated. Gemini Flash 3 showed no comparable asymmetry, so both truth arms were retained; its answer logits were read with the reasoning-cut procedure described in the Models section.

\subsubsection*{Count-comparison task}
Candidate questions were those whose program terminated in \texttt{equal\_integer} (whether two object counts are equal), restricted to one or two relational operations so that the counted sets were defined relative to other objects in the scene. We kept questions with at least one non-zero queried count and removed duplicate image--question pairs. Qwen2.5-VL was run on $1{,}000$ yes-truth questions at $\sigma = 0, 2, 4$, a fine grid from $4$ to $10$ in steps of $0.25$, and coarser levels at $12, 14, 16, 18, 20, 25, 30$; Gemini Flash 3 used the D/N letter pair on $500$ questions per truth arm at $\sigma = 0$--$10$ in unit steps.

\subsubsection*{Simple-existence task}
Candidate questions were those terminating in \texttt{exist}, restricted to simple two-attribute filter-only programs (two \texttt{filter\_*} operations, e.g.\ colour and shape; questions with relational or \texttt{same\_*} operations were excluded). Both models used the M/V letter pair. Qwen2.5-VL was run on $800$ yes-truth questions at $\sigma = 0, 1$ then a fine grid from $2$ to $10$ in steps of $0.25$; Gemini Flash 3 was evaluated on both truth arms over the same grid.

\paragraph{Difficulty controls.}
To test whether answer-logit confidence merely proxied visual or question difficulty, we augmented each trial with scene and question metadata from the CLEVR annotations: scene-level controls (total objects; unique colours, shapes, sizes, and materials; maximum objects sharing an attribute; maximum identical object types; and spatial-crowding measures from pairwise object distances) and, where applicable, program controls (program length and counts of filter, relation, same-attribute, count, existence, union, and intersection operations), plus the queried counts and their absolute difference for count questions. After removing highly collinear controls, we tested whether $|\mathrm{LD}|$ improved prediction beyond blur strength and these controls (nested likelihood-ratio tests) and whether it predicted correctness within coarse blur bins after including them (fixed-blur-bin analyses).

\subsection*{Models}
\paragraph{Non-reasoning models.}
We evaluated three non-reasoning multimodal language models, all run locally with the HuggingFace \texttt{transformers} library: Qwen2.5-VL-7B-Instruct (\texttt{Qwen/Qwen2.5-VL-7B-Instruct}), Gemma~3 12B (\texttt{google/gemma-3-12b-it}), and Gemma~4 12B (\texttt{google/gemma-4-12B-it}). Each was prompted to answer with a single letter or word (see the task descriptions), so the model returned a direct answer with no reasoning trace and the answer logits could be read from the first response token. Although Gemma~4 is reasoning-capable, this single-token response format elicits a direct answer, placing it in the same non-reasoning regime as the other two models.

\paragraph{Reasoning model.}
We evaluated one reasoning model, Gemini Flash~3 preview, accessed through the Beyond API on Google infrastructure. Confidence was read using the reasoning-cut procedure below.

\paragraph{Reasoning-cut readout (Gemini Flash~3).}
A reasoning model does not expose a well-defined answer logit until it commits to an answer, and the signal saturates once the answer is verbalized. On each trial the model was prompted to reason step by step and to end with a one-letter answer; the generated trace was truncated at $90\%$ of its length, and the model was then asked to answer the original question from the reasoning produced so far, yielding a graded answer-logit difference before saturation. We applied this $90\%$ reasoning-cut readout to Gemini Flash~3 on both the contrast discrimination task and the CLEVR count and existence tasks. At the $90\%$ cut, accuracy did not differ significantly from accuracy after the complete reasoning trace ($p>0.1$).
\clearpage
\bibliographystyle{plainnat}  
\bibliography{references_SDC}

\clearpage
\section*{Appendix}
\subsection*{Supplemental Results: Gemma 3 12B}
\begin{figure}[!t]
\centering
\includegraphics[angle = -90, width=0.8\textwidth]{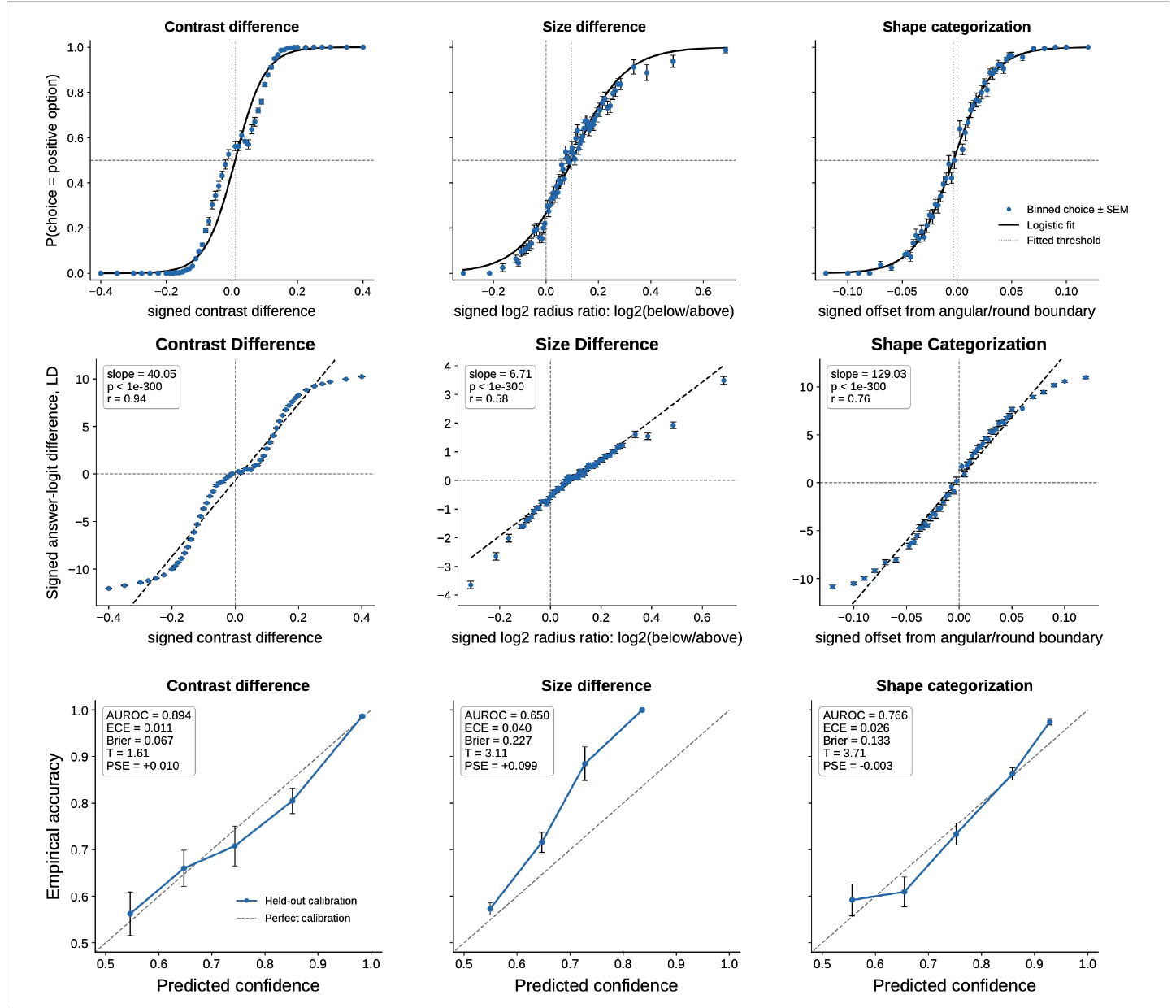}
\caption{\textbf{Gemma 3 12B: Choice behaviour, LD-stimulus strength plot and calibration across visual decision tasks.}
Top: Psychometric curves for the three perceptual tasks. Points show the empirical probability of choosing the positive-coded option, with SEM error bars; solid lines show fitted logistic psychometric curves. The signed stimulus axis is defined so that positive values favour the positive-coded response: higher contrast/right patch for contrast, lower circle for size, and round for shape categorization. Gemma 3 12B showed graded stimulus--choice relationships in all tasks, but performance was substantially weaker in the size task. In that task, the model showed a pronounced bias against the BELOW response: the fitted point of subjective equality was shifted to approximately $+0.099$ signed-size units, and at the objective boundary the model was predicted to choose BELOW on only approximately $27\%$ of trials. Bottom: calibration plots.}
\label{fig:gemma3_psychom}
\end{figure}

\begin{figure}[!t]
    \centering
    \includegraphics[angle=-90, width=1\textwidth]{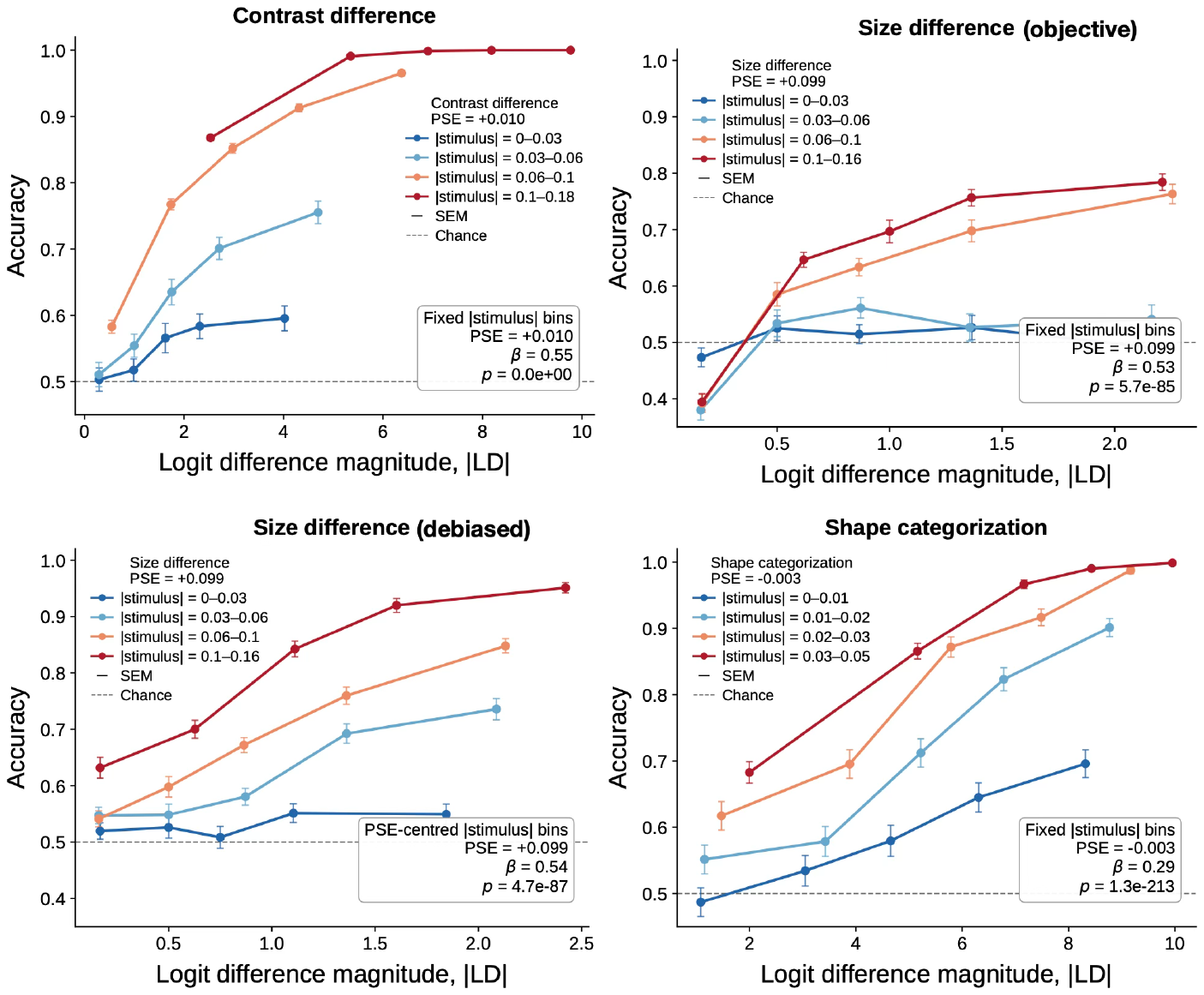}
\caption{\textbf{Answer-logit confidence predicts correctness within fixed stimulus-strength bins in Gemma 3 12B.}
Trials were grouped into coarse stimulus-strength bins and then binned by logit-difference magnitude within each strength bin. Points show empirical accuracy in each logit-difference bin, connected separately for each fixed stimulus-strength range. Insets report pooled logistic regressions predicting correctness from logit-difference magnitude while controlling for fixed stimulus-strength bin, $\mathrm{correct} \sim C(|\mathrm{stimulus}|, \mathrm{bin}) + |LD|$. For contrast and shape, and for the \emph{top-right} size panel, stimulus bins and correctness are defined objectively, as in the Qwen analysis. However, Gemma 3 12B showed a pronounced bias in the size task. We therefore additionally show a \emph{model-relative} size analysis (\emph{bottom-left}), in which trials are grouped by distance from the model's fitted point of subjective equality and correctness is defined relative to the side of the model's own subjective boundary. This model-relative analysis asks whether answer-logit confidence tracks accuracy within the model's internal decision space, rather than only with respect to the externally defined task boundary. Positive coefficients indicate that trial-to-trial variation in answer-logit confidence predicts correctness beyond coarse stimulus-strength matching. Error bars are SEM.}
\label{fig:fixed_ss_gemma3}
\end{figure}

\begin{figure}[!t]
    \centering
    \includegraphics[angle = -90, width=1\textwidth]{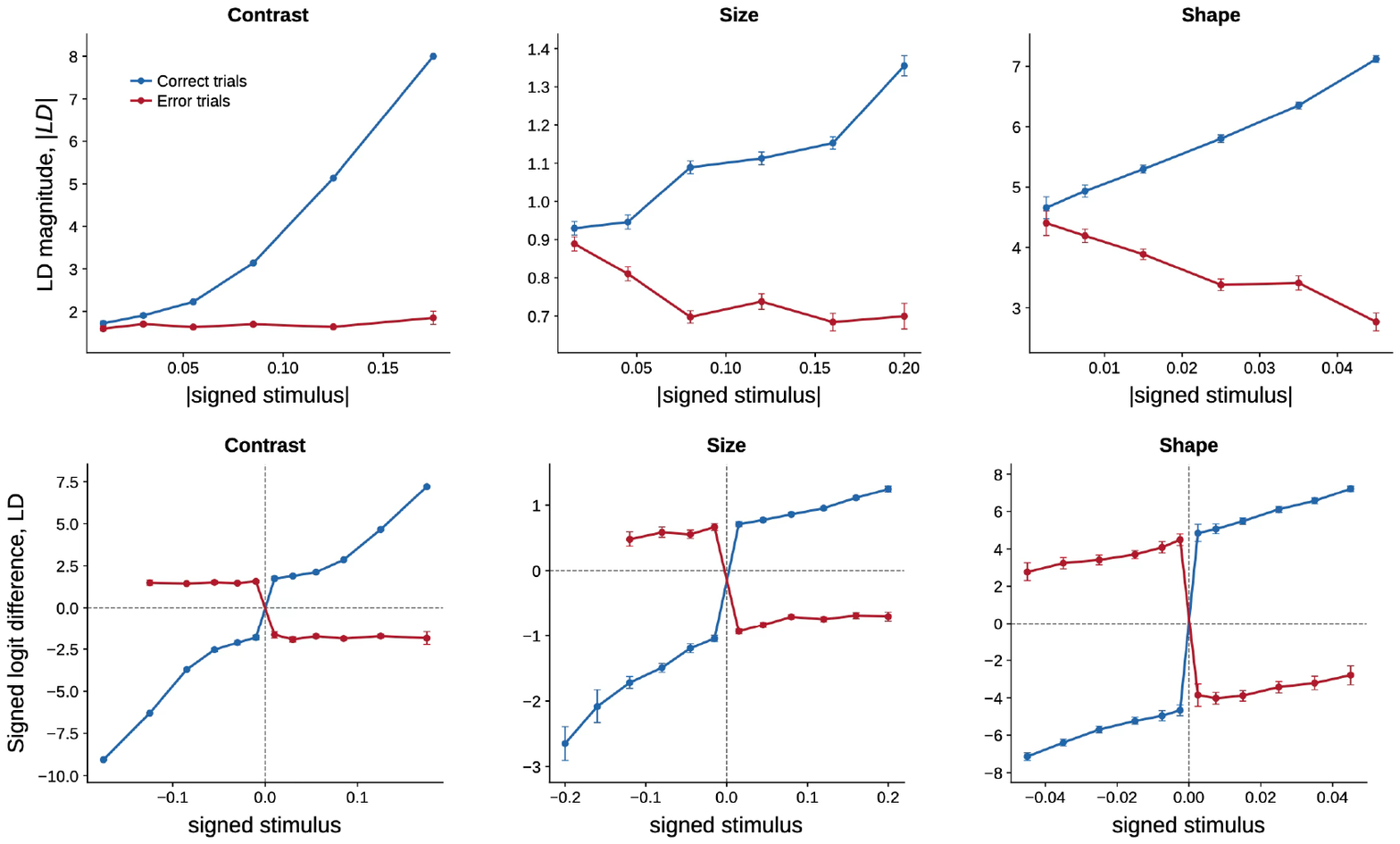}
\caption{\textbf{Gemma 3 12B: Decision-strength and signed evidence structure in answer logits.}
Top: Logit-difference magnitude, \(|LD|\), plotted against absolute stimulus strength, \(|\mathrm{signed}|\). Correct trials show increasing \(|LD|\) with stimulus strength, whereas error trials remain lower and occur mainly near the decision boundary. Points show binned means with bootstrap 95\% confidence intervals. Bottom: Full signed X-pattern. The task-specific answer-logit difference \(LD\) is plotted against the signed stimulus variable, separately for correct and error trials. Correct trials lie on the stimulus-congruent side of \(LD=0\), whereas error trials lie on the opposite side. }
\label{fig:Gemma_X_composite}
\end{figure}

\begin{figure}[!t]
\centering
\includegraphics[angle=-90, width=1\textwidth]{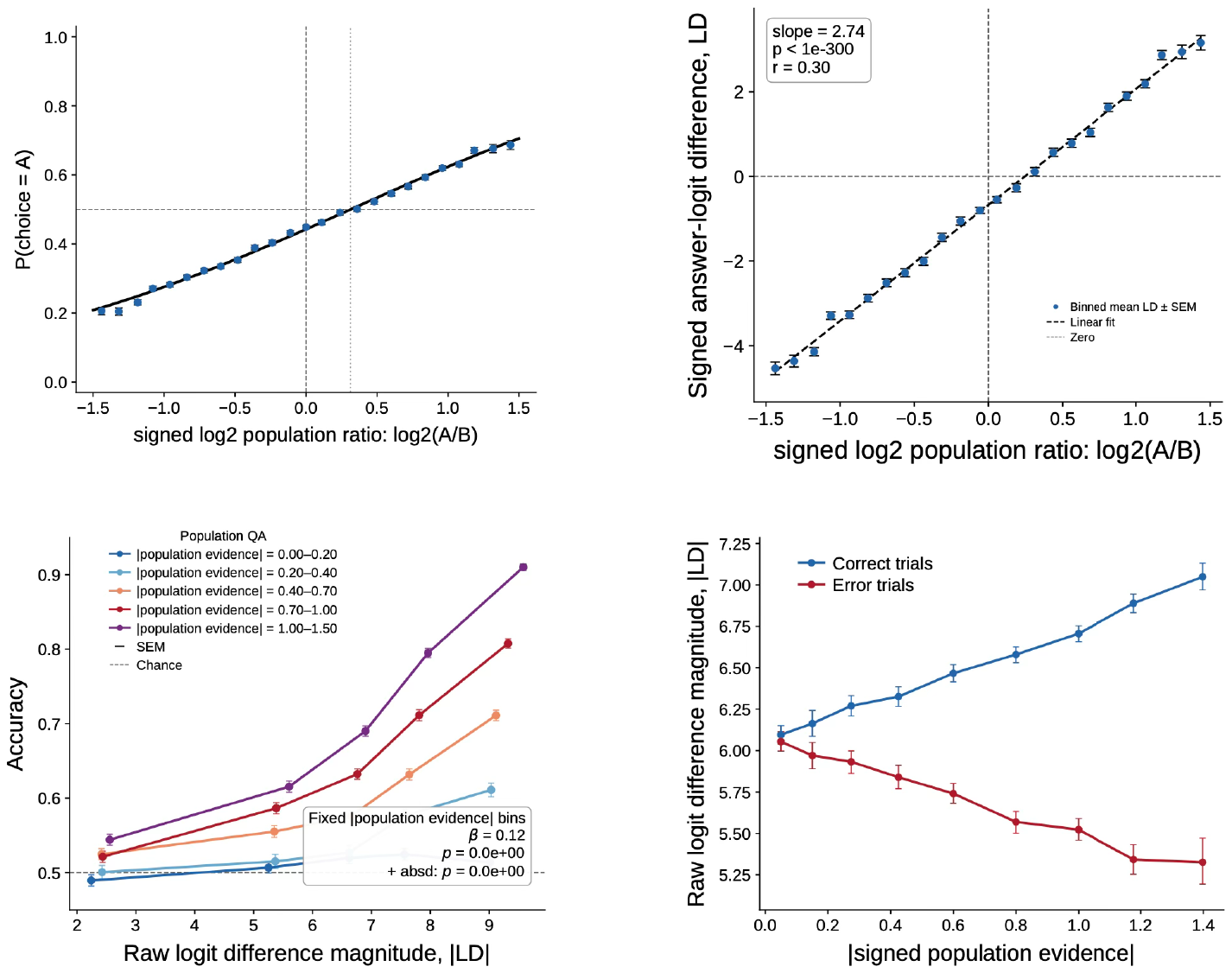}
\caption{\textbf{Gemma 3 12B population QA: Results} Top left: psychometric curve for population comparison trials. The signed stimulus is the $\log_2$ population ratio, $\log_2(A/B)$, so positive values favour answer A.  Top right: signed LD  is correlated with binned stimulus strength. Bottom left: fixed-evidence analysis; trials were grouped by absolute population evidence and then binned by raw logit-difference magnitude. Bottom right: correct/error X-pattern analysis, plotting raw logit-difference magnitude against absolute population evidence. Error bars show SEM for psychometric, signed LD, and fixed-bin accuracy panels, and bootstrap 95\% confidence intervals for the correct/error LD panel.}
\label{fig:gemma3_population}
\end{figure}

Gemma 3 12B reproduced the encoding and readout signatures across the three perceptual tasks (Figures~\ref{fig:gemma3_psychom},~\ref{fig:fixed_ss_gemma3},~\ref{fig:Gemma_X_composite}; Table~\ref{tab:gemma3_12b_summary}). Two task-specific points: the size task carried a strong response bias, so we additionally confirmed the fixed-strength result in a model-relative analysis defining correctness relative to the model's own PSE (Figure~\ref{fig:fixed_ss_gemma3}); and the contrast error branch was flat rather than falling (Table~\ref{tab:gemma3_12b_summary}(D)), the only departure from the full folded-X. We treat the flat branch as neutral: it is neither the falling branch predicted by SDC nor the rising branch sometimes observed when confidence fails to follow the SDC geometry. LD nevertheless predicted correctness within fixed stimulus-strength bins, with $|LD|$ predicting correctness (Table~\ref{tab:gemma3_12b_summary}(C)). The signatures also held in the population QA task, with a clean folded-X carrying a strongly negative error branch despite a shallow psychometric (Figure~\ref{fig:gemma3_population}; Table~\ref{tab:gemma3_12b_population_summary}).

\subsection*{Supplemental Results: Gemma 4 12B}
\begin{figure}[!t]
    \centering
    \includegraphics[angle = -90, width=1\textwidth]{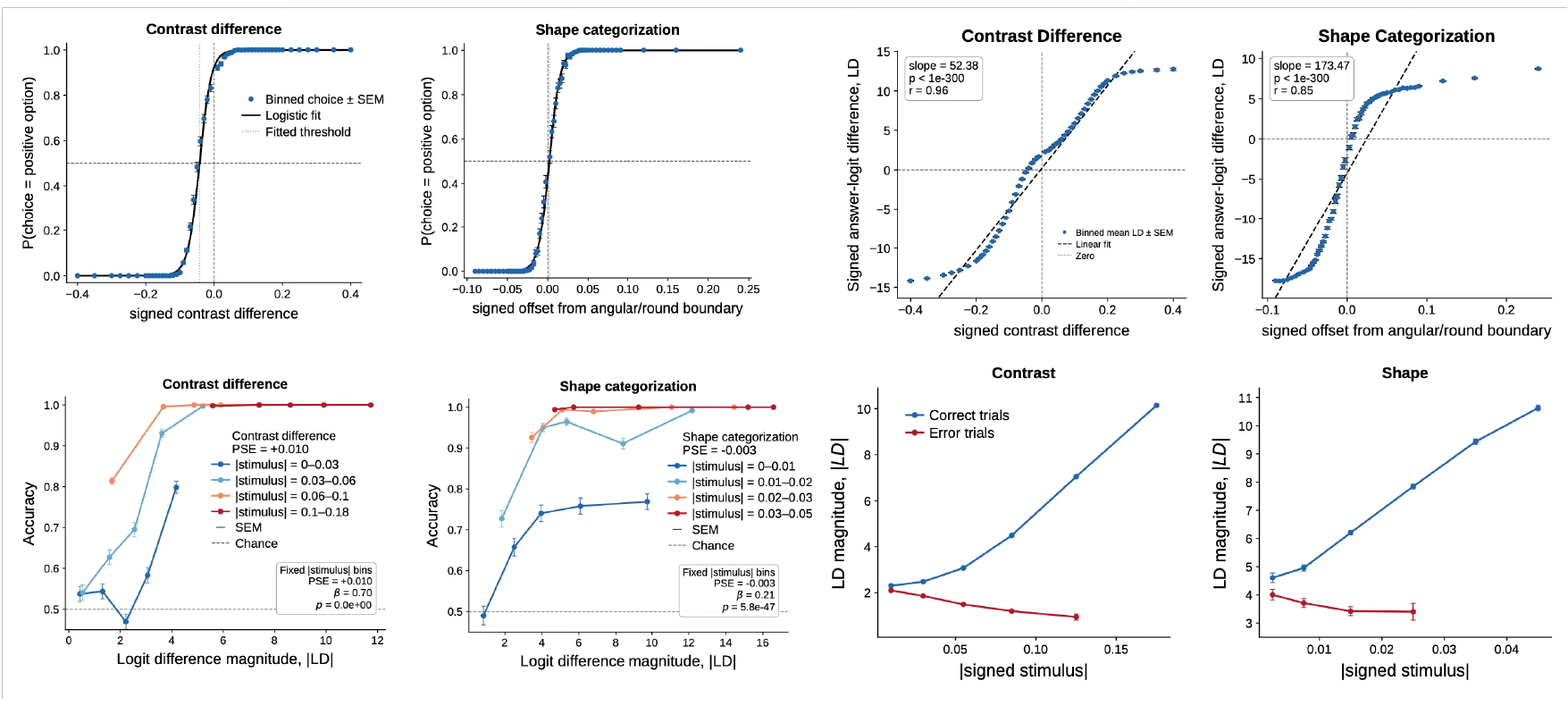}
\caption{\textbf{Perceptual tasks and Gemma 4 12B}
Top: psychometric curves for contrast discrimination and shape categorization. Points show binned choice probabilities with SEM error bars; solid lines show fitted logistic psychometric curves and dotted vertical lines mark fitted thresholds. Middle: fixed stimulus-strength analyses. Trials were grouped by objective stimulus strength and then binned by answer-logit confidence, $|LD|$; points show empirical accuracy with SEM error bars, and insets report pooled logistic regressions controlling for stimulus-strength bin. Middle: plot of signed LD against stimulus strength. Bottom: X-pattern analyses. Mean $|LD|$ is plotted as a function of objective stimulus strength, separately for correct and error trials; error bars show SEM. The size task is omitted because objective performance was near chance. Full quantitative results are summarized in Table~\ref{tab:gemma4_12b_contrast_shape_summary}.}
\label{fig:Gemma4_composite}
\end{figure}

Gemma 4 12B reproduced the signatures on contrast and shape, including the full folded-X (Figure~\ref{fig:Gemma4_composite}; Table~\ref{tab:gemma4_12b_contrast_shape_summary}). The size task was excluded for near-chance objective performance (accuracy $0.509$), driven by a large response bias rather than graded uncertainty.

\begin{figure}[!t]
    \centering
    \includegraphics[width=1\textwidth]{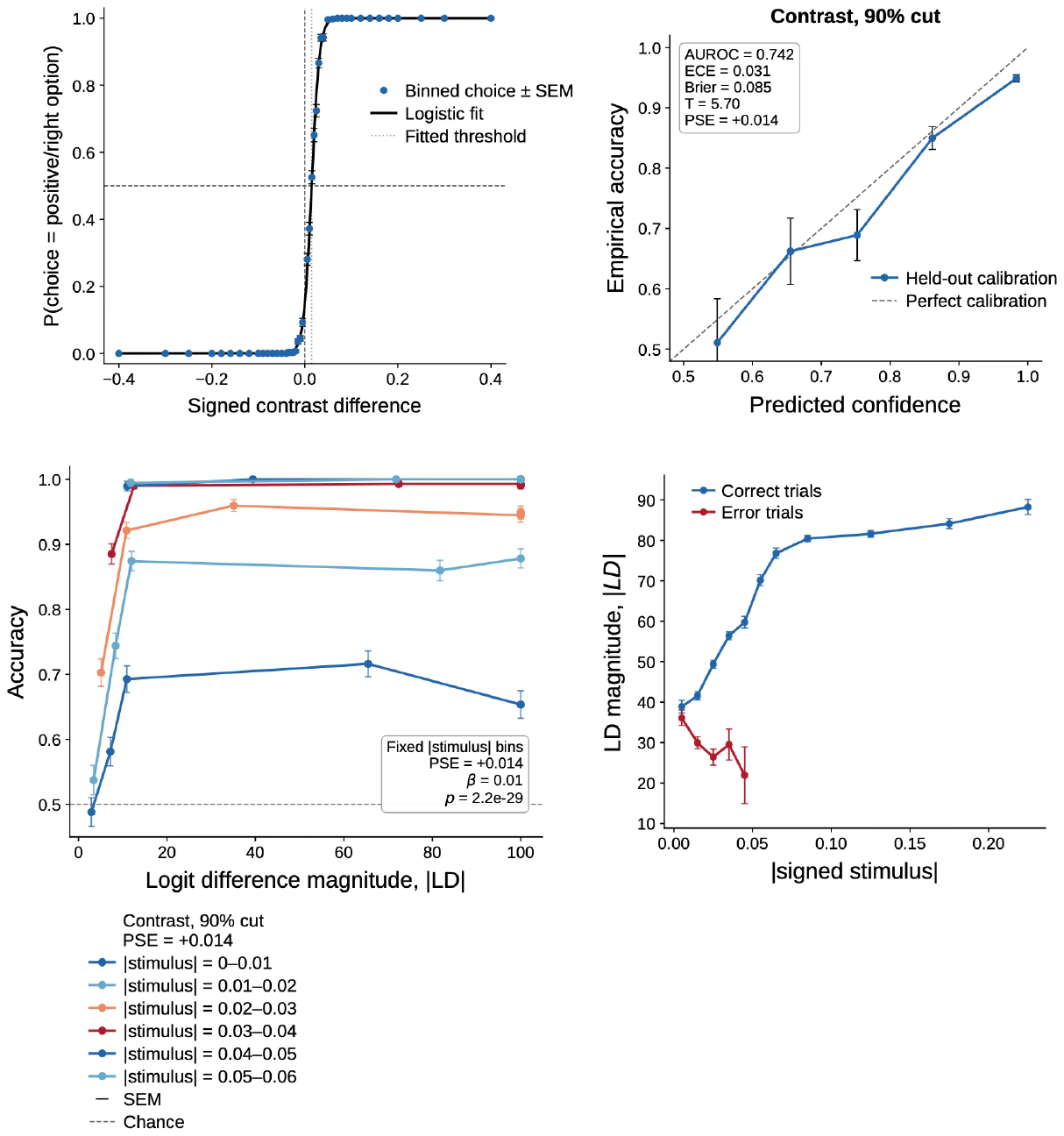}
\caption{\textbf{Contrast Task and Gemini Flash 3 -- a reasoning model}
Top left: psychometric curves for contrast discrimination. Points show binned choice probabilities with SEM error bars; solid lines show fitted logistic psychometric curves and dotted vertical lines mark fitted thresholds.Top right: calibration plot. Bottom left: fixed stimulus-strength analyses. Trials were grouped by objective stimulus strength and then binned by answer-logit confidence, $|LD|$; points show empirical accuracy with SEM error bars, and insets report pooled logistic regressions controlling for stimulus-strength bin. Bottom right: X-pattern analyses. Mean $|LD|$ is plotted as a function of objective stimulus strength, separately for correct and error trials; error bars show SEM. Quantitative results are summarized in Table~\ref{tab:fiercefalcon_contrast_90}.}
\label{fig:Fierce_Falcon_composite}
\end{figure}

\paragraph{Gemini Flash 3 (reasoning model).}
On the contrast task at the 90\% reasoning cut, Gemini Flash 3 showed the same signatures---a steep psychometric, fixed-strength prediction of correctness, and a clear folded-X (Figure~\ref{fig:Fierce_Falcon_composite}; Table~\ref{tab:fiercefalcon_contrast_90}).

\subsection*{Supplemental Discussion}
\paragraph{Recognition-memory example.}
Consider a recognition-memory task in which the experimenter defines the relevant evidence as the difference in familiarity acquired during the experiment,

\[
\Delta_{\rm expt}.
\]

Suppose, however, that the model cannot distinguish experimentally acquired familiarity from pre-existing real-world familiarity,

\[
\Delta_{\rm rw},
\]

so that its latent decision variable is

\[
d
=
\Delta_{\rm expt}
+
\Delta_{\rm rw}
+
\epsilon.
\]

The experimenter-defined correctness depends only on $\Delta_{\rm expt}$, but the observed decision variable depends on both sources of familiarity. Consequently, the normative posterior probability that the current choice is experimentally correct would be

\[
P(C_{\rm expt}\mid d)
=
\frac{
\sum_{\Delta_{\rm rw}}
p(d\mid \Delta_{\rm expt}>0,\Delta_{\rm rw})
P(\Delta_{\rm expt}>0,\Delta_{\rm rw})
}{
\sum_{\Delta_{\rm expt}}
\sum_{\Delta_{\rm rw}}
p(d\mid \Delta_{\rm expt},\Delta_{\rm rw})
P(\Delta_{\rm expt},\Delta_{\rm rw})
},
\]

rather than the simpler expression obtained under the experimenter-defined model alone. That is, because $\Delta_{\rm rw}$ also contributes to the generation of $d$, the posterior defined over the experimenter's evidence axis no longer coincides with the posterior defined over the model's effective evidence variable.

If the experimenter analyses behaviour only with respect to $\Delta_{\rm expt}$, the fixed-strength and folded-X signatures need not hold, since trial-to-trial variation in the decision variable now reflects both experimental and real-world familiarity. This does not by itself imply that the model departs from the normative model defined over its own effective evidence variable; it illustrates how the experimenter-defined evidence axis may differ from the model's effective evidence variable.

\begin{figure}[!t]
\centering
\includegraphics[angle=-90, width=1\textwidth]{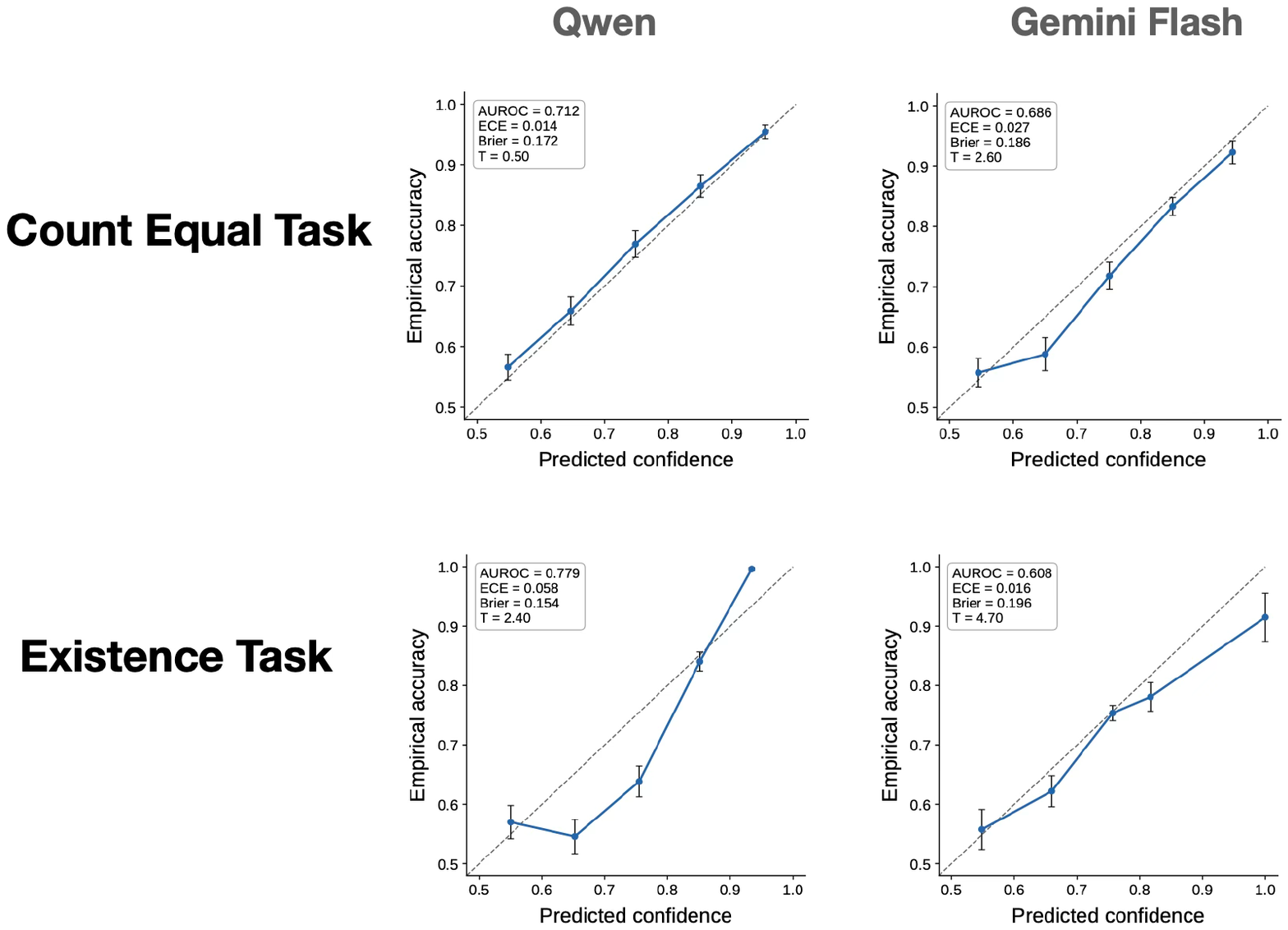}
\caption{\textbf{Calibrated confidence in CLEVR for Qwen 2.5 7B and Gemini Flash 3}}
\label{fig:CLEVR_calibration}
\end{figure}

\begin{table}[htbp]

\small                       
\setlength{\tabcolsep}{1pt}
\caption{\textbf{Qwen 2.5 7B: confidence summary across perceptual tasks.}
Panels preserve all reported quantities. Accuracy and PSE are shown with 95\% CIs; logistic slopes as estimate $\pm$ SE.}
\label{tab:qwen_perceptual_summary}

\vspace{0.5em}
\textbf{(A) Psychophysical performance.}\par\smallskip
\makebox[\textwidth][c]{%
\begin{tabular}{lrrrrrrl}

\toprule
Task & $N$ & Accuracy & Logistic slope & PSE & 25--75 width & pseudo-$R^2$ & Bias \\
\midrule
Contrast difference
& 63{,}500 & $0.922\,[0.920,\,0.924]$ & $25.01 \pm 0.19$ & $+0.0193\,[+0.0180,\,+0.0206]$ & $0.0879$ & $0.714$ & Against higher/right \\
Size difference
& 11{,}640 & $0.806\,[0.799,\,0.813]$ & $24.36 \pm 0.47$ & $+0.0797\,[+0.0772,\,+0.0822]$ & $0.0902$ & $0.572$ & Against below \\
Shape categorization
& 10{,}900 & $0.868\,[0.862,\,0.875]$ & $105.82 \pm 2.03$ & $+0.0043\,[+0.0037,\,+0.0049]$ & $0.0208$ & $0.554$ & Against round \\
\bottomrule

\end{tabular}}

\vspace{1em}
\textbf{(B) Confidence predicts correctness within fixed stimulus-strength bins.}\par\smallskip
\begin{tabular}{llrrrrrrrr}
\toprule
Task & Predictor & $N$ & Bins & Accuracy & $\beta$ & SE & $z$ & $p$ & Odds ratio \\
\midrule
Contrast difference
& Answer-logit confidence & 48{,}000 & 4 & $0.898$ & $4.117$ & $0.077$ & $53.42$ & $<10^{-300}$ & $61.37$ \\
Size difference
& Answer-logit confidence & 8{,}320 & 4 & $0.739$ & $1.462$ & $0.045$ & $32.29$ & $9.7\times10^{-229}$ & $4.32$ \\
Shape categorization
& Answer-logit confidence & 9{,}500 & 4 & $0.849$ & $1.968$ & $0.101$ & $19.47$ & $1.9\times10^{-84}$ & $7.16$ \\
\bottomrule
\end{tabular}

\vspace{1em}
\textbf{(C) X-pattern regression} ($|\mathrm{LD}| \sim |\mathrm{stimulus}| \times \mathrm{Correct}$; robust HC3 SEs)\par\smallskip
\begin{tabular}{lrrrrrr}
\toprule
Task & $N$ & Error slope & $p_{\mathrm{error}}$ & Correct slope & $p_{\mathrm{correct}}$ & Interaction ($p_{\mathrm{int}}$) \\
\midrule
Contrast difference
& 51{,}600 & $-1.108 \pm 0.080$ & $<10^{-40}$ & $+8.076 \pm 0.051$ & $<10^{-300}$ & $+9.185 \pm 0.095$ ($<10^{-300}$) \\
Size difference
& 10{,}660 & $-3.701 \pm 0.195$ & $<10^{-75}$ & $+7.209 \pm 0.172$ & $<10^{-300}$ & $+10.910 \pm 0.260$ ($<10^{-300}$) \\
Shape categorization
& 9{,}150 & $-10.617 \pm 1.151$ & $<10^{-19}$ & $+28.508 \pm 0.674$ & $<10^{-300}$ & $+39.125 \pm 1.334$ ($<10^{-180}$) \\
\bottomrule
\end{tabular}
\vspace{0.5ex}

\end{table}
\begin{table}[htbp]
\centering
\small
\setlength{\tabcolsep}{4pt}
\caption{\textbf{Gemma 3 12B: confidence summary across perceptual tasks.}
All three tasks are retained; the size task was above chance but showed a pronounced response bias. Panels mirror the per-analysis tables and will be consolidated in the final version.}
\label{tab:gemma3_12b_summary}

\vspace{0.5em}
\noindent\textbf{(A) Psychophysical performance.}\par\smallskip
\makebox[\textwidth][c]{%
\begin{tabular}{lrrrrrrl}
\toprule
Task & $N$ & Accuracy & Logistic slope & PSE & 25--75 width & pseudo-$R^2$ & Bias \\
\midrule
Contrast difference
& 63{,}500 & $0.903\,[0.901,\,0.906]$ & $22.13 \pm 0.16$ & $+0.0097\,[+0.0083,\,+0.0111]$ & $0.0993$ & $0.667$ & Against higher/right \\
Size difference
& 18{,}960 & $0.611\,[0.604,\,0.618]$ & $10.23 \pm 0.19$ & $+0.0990\,[+0.0959,\,+0.1021]$ & $0.2148$ & $0.146$ & Against below \\
Shape categorization
& 12{,}820 & $0.800\,[0.794,\,0.807]$ & $55.77 \pm 0.96$ & $-0.0033\,[-0.0041,\,-0.0025]$ & $0.0394$ & $0.369$ & Toward round \\
\bottomrule
\end{tabular}}

\vspace{1em}
\noindent\textbf{(B) Answer-logit confidence predicts correctness beyond stimulus strength.}\par\smallskip
\makebox[\textwidth][c]{%
\begin{tabular}{lrrrrrrrr}
\toprule
Task & $N$ & Accuracy & AUROC$_{\mathrm{stim}}$ & AUROC$_{\mathrm{LD}}$ & AUROC$_{\mathrm{both}}$ & $\Delta$AUROC & LR & $p$ \\
\midrule
Contrast difference
& 63{,}500 & $0.903$ & $0.881$ & $0.900$ & $0.914$ & $+0.033$ & $3643.9$ & $<10^{-300}$ \\
Size difference
& 18{,}960 & $0.611$ & $0.612$ & $0.664$ & $0.675$ & $+0.063$ & $1042.0$ & $1.34\times10^{-228}$ \\
Shape categorization
& 12{,}820 & $0.800$ & $0.740$ & $0.771$ & $0.810$ & $+0.071$ & $1054.4$ & $2.71\times10^{-231}$ \\
\bottomrule
\end{tabular}}

\vspace{1em}
\noindent\textbf{(C) Confidence predicts correctness within fixed stimulus-strength bins.}\par\smallskip
\makebox[\textwidth][c]{%
\begin{tabular}{llrrrrrrrr}
\toprule
Task & Predictor & $N$ & Bins & Accuracy & $\beta$ & SE & $z$ & $p$ & Odds ratio \\
\midrule
Contrast difference
& Answer-logit confidence & 48{,}000 & 4 & $0.872$ & $0.546$ & $0.010$ & $52.11$ & $<10^{-300}$ & $1.73$ \\
Size difference
& Answer-logit confidence & 14{,}450 & 4 & $0.568$ & $0.532$ & $0.027$ & $19.53$ & $5.74\times10^{-85}$ & $1.70$ \\
Shape categorization
& Answer-logit confidence & 11{,}320 & 4 & $0.775$ & $0.286$ & $0.009$ & $31.19$ & $1.34\times10^{-213}$ & $1.33$ \\
\bottomrule
\end{tabular}}

\vspace{1em}
\noindent\textbf{(D) X-pattern regression} ($|\mathrm{LD}| \sim |\mathrm{stimulus}| \times \mathrm{Correct}$; robust HC3 SEs).\par\smallskip
\makebox[\textwidth][c]{%
\begin{tabular}{lrrrrrr}
\toprule
Task & $N$ & Error slope & $p_{\mathrm{error}}$ & Correct slope & $p_{\mathrm{correct}}$ & Interaction ($p_{\mathrm{int}}$) \\
\midrule
Contrast difference
& 51{,}600 & $+0.287 \pm 0.500$ & $0.566$ & $+48.738 \pm 0.196$ & $<10^{-300}$ & $+48.451 \pm 0.537$ ($<10^{-300}$) \\
Size difference
& 17{,}020 & $-1.132 \pm 0.152$ & $<10^{-13}$ & $+2.080 \pm 0.130$ & $<10^{-55}$ & $+3.212 \pm 0.200$ ($<10^{-55}$) \\
Shape categorization
& 11{,}160 & $-34.622 \pm 3.830$ & $<10^{-18}$ & $+58.765 \pm 2.174$ & $<10^{-150}$ & $+93.387 \pm 4.404$ ($<10^{-95}$) \\
\bottomrule
\end{tabular}}
\end{table}
\begin{table}[t]
\centering
\caption{\textbf{Abstention is strongly predicted by answer-logit confidence.}
The model abstained on 87.7\% of trials and achieved near-perfect accuracy on committed responses. AUROC values quantify prediction of abstention. Nested logistic models compare physical stimulus strength and answer-logit confidence as predictors of abstention.}
\label{tab:abstention_summary}
\begin{tabular}{lr}
\toprule
Metric & Value \\
\midrule
Abstention rate & 0.877 \\
Commit rate & 0.123 \\
Forced-choice accuracy & 0.922 \\
Accuracy given commit & 0.9995 \\
Accuracy given abstain & 0.912 \\
\midrule
AUROC (stimulus strength $\rightarrow$ abstention) & 0.796 \\
AUROC (answer-logit confidence $\rightarrow$ abstention) & 0.840 \\
AUROC (combined model) & 0.843 \\
\midrule
Nested-model LR: LD adds beyond stimulus strength & 2249.5 \\
Nested-model coefficient for low LD ($z$-scored) & 0.935 \\
Nested-model coefficient for difficulty ($z$-scored) & 0.220 \\
\midrule
Within-bin coefficient for low LD & 0.867 \\
Within-bin odds ratio & 2.38 \\
Within-bin LR statistic & 2748.0 \\
\midrule
Overlap with optimal LD-threshold policy & 0.916 \\
Committed accuracy gain of optimal policy & 0.0005 \\
\bottomrule
\end{tabular}
\end{table}
\begin{table}[htbp]
\centering
\small
\caption{\textbf{Qwen 2.5 7B, Population QA (city-population comparison): consolidated results.}
Calibration uses raw $|LD|$ on held-out trials after temperature scaling.}
\label{tab:qwen_population_summary}

\vspace{0.5em}
\textbf{(A) Psychometric and calibration summary.}
\begin{tabular}{lrrrrrrrr}
\toprule
Task & $N$ & Accuracy & $P(A)$ & Intercept & Slope & PSE & pseudo-$R^2$ & AUROC$_{|LD|}$ \\
\midrule
Population QA
& 50{,}000 & $0.618$ & $0.654$ & $0.708$ & $0.900$ & $-0.787$ & $0.069$ & $0.590$ \\
\bottomrule
\end{tabular}

\vspace{1em}
\textbf{(B) Raw logit magnitude predicts correctness beyond population evidence.}
\begin{tabular}{lrrrrrrr}
\toprule
Model & $N$ & Accuracy & AUROC$_x$ & $\beta_x$ & SE & $z$ & OR \quad($p$) \\
\midrule
Fixed evidence bins
& 50{,}000 & $0.618$ & $0.590$ & $0.306$ & $0.011$ & $28.65$ & $1.36$ \;($1.2\times10^{-207}$) \\
\bottomrule
\end{tabular}

\vspace{1em}
\textbf{(C) X-pattern regression} (raw $|\mathrm{LD}|$ on absolute population evidence, by correctness).
\begin{tabular}{lrrr}
\toprule
Term & Estimate & SE & $t$ \quad($p$) \\
\midrule
Correct-trial slope & $+0.304$ & $0.015$ & $20.3$ \;($1.6\times10^{-94}$) \\
Error-trial slope   & $-0.145$ & $0.016$ & $-9.1$ \;($9.2\times10^{-19}$) \\
Interaction         & $+0.449$ & $0.022$ & $20.4$ \;($3.9\times10^{-92}$) \\
\bottomrule
\end{tabular}
\end{table}

\begin{table}[htbp]
\centering
\small

\caption{\textbf{Gemma 4 12B: confidence summary for contrast and shape.}
The size task is omitted because objective performance was near chance ($0.509$), dominated by a response bias rather than graded uncertainty. Panels mirror the per-analysis Gemma 3 12B tables and will be consolidated in the final version.}
\label{tab:gemma4_12b_contrast_shape_summary}

\vspace{0.5em}
\textbf{(A) Psychophysical performance.}

\makebox[\textwidth][c]{%
\begin{tabular}{lrrrrrrl}
\toprule
Task & $N$ & Accuracy & Logistic slope & PSE & 25--75 width & pseudo-$R^2$ & Bias \\
\midrule
Contrast difference
& 63{,}500 & $0.955\,[0.954,\,0.957]$ & $58.44 \pm 0.81$ & $-0.0426\,[-0.0436,\,-0.0416]$ & $0.0376$ & $0.900$ & Toward higher/right \\
Shape categorization
& 14{,}520 & $0.929\,[0.925,\,0.933]$ & $150.60 \pm 2.97$ & $+0.0019\,[+0.0014,\,+0.0024]$ & $0.0146$ & $0.760$ & Against round \\
\bottomrule
\end{tabular}}

\vspace{1em}
\textbf{(B) Answer-logit confidence predicts correctness beyond stimulus strength.}

\makebox[\textwidth][c]{%
\begin{tabular}{lrrrrrrrr}
\toprule
Task & $N$ & Accuracy & AUROC$_{\mathrm{stim}}$ & AUROC$_{\mathrm{LD}}$ & AUROC$_{\mathrm{both}}$ & $\Delta$AUROC & LR & $p$ \\
\midrule
Contrast difference
& 63{,}500 & $0.955$ & $0.952$ & $0.958$ & $0.970$ & $+0.018$ & $2118.1$ & $<10^{-300}$ \\
Shape categorization
& 14{,}520 & $0.929$ & $0.908$ & $0.815$ & $0.922$ & $+0.014$ & $268.1$ & $2.99\times10^{-60}$ \\
\bottomrule
\end{tabular}}

\vspace{1em}
\textbf{(C) Confidence predicts correctness within fixed stimulus-strength bins.}

\makebox[\textwidth][c]{%
\begin{tabular}{llrrrrrrrr}
\toprule
Task & Predictor & $N$ & Bins & Accuracy & $\beta$ & SE & $z$ & $p$ & Odds ratio \\
\midrule
Contrast difference
& Answer-logit confidence & 48{,}000 & 4 & $0.941$ & $0.701$ & $0.017$ & $41.96$ & $<10^{-300}$ & $2.02$ \\
Shape categorization
& Answer-logit confidence & 12{,}000 & 4 & $0.914$ & $0.209$ & $0.015$ & $14.39$ & $5.8\times10^{-47}$ & $1.23$ \\
\bottomrule
\end{tabular}}

\vspace{1em}
\textbf{(D) X-pattern regression} ($|\mathrm{LD}| \sim |\mathrm{stimulus}| \times \mathrm{Correct}$; robust HC3 SEs).

\makebox[\textwidth][c]{%
\begin{tabular}{lrrrrrr}
\toprule
Task & $N$ & Error slope & $p_{\mathrm{error}}$ & Correct slope & $p_{\mathrm{correct}}$ & Interaction ($p_{\mathrm{int}}$) \\
\midrule
Contrast difference
& 51{,}600 & $-12.80 \pm 0.75$ & $<10^{-60}$ & $+58.51 \pm 0.18$ & $<10^{-300}$ & $+71.31 \pm 0.78$ ($<10^{-300}$) \\
Shape categorization
& 11{,}400 & $-34.55 \pm 13.01$ & $7.9\times10^{-3}$ & $+150.94 \pm 3.20$ & $<10^{-300}$ & $+185.49 \pm 13.40$ ($<10^{-40}$) \\
\bottomrule
\end{tabular}}
\end{table}

\begin{table}[htbp]
\centering
\small
\caption{\textbf{Gemma 3 12B, Population QA: consolidated results.}
AB/BA-balanced 100k dataset (original and swapped A/B presentations concatenated). The signed stimulus variable is the $\log_2$ population ratio of the positive-coded option. Panels will be consolidated in the final version.}
\label{tab:gemma3_12b_population_summary}

\vspace{0.5em}
\textbf{(A) Psychometric and calibration summary.}
\begin{tabular}{lrrrrrrrr}
\toprule
Task & $N$ & Accuracy & $P(A)$ & Intercept & Slope & PSE & pseudo-$R^2$ & AUROC$_{|LD|}$ \\
\midrule
Population QA
& 100{,}000 & $0.605$ & $0.446$ & $-0.231$ & $0.737$ & $+0.313$ & $0.049$ & $0.592$ \\
\bottomrule
\end{tabular}

\vspace{1em}
\textbf{(B) Raw logit magnitude predicts correctness beyond population evidence.}
\begin{tabular}{lrrrrrrrr}
\toprule
Model & $N$ & Accuracy & AUROC$_x$ & $\beta_x$ & SE & $z$ & OR & $p$ \\
\midrule
Fixed evidence bins
& 100{,}000 & $0.605$ & $0.592$ & $0.1218$ & $0.0027$ & $45.19$ & $1.129$ & $<10^{-300}$ \\
Fixed bins $+$ continuous $|\Delta|$
& 100{,}000 & $0.605$ & $0.592$ & $0.1214$ & $0.0027$ & $45.00$ & $1.129$ & $<10^{-300}$ \\
\bottomrule
\end{tabular}

\vspace{1em}
\textbf{(C) X-pattern regression} (raw $|\mathrm{LD}|$ on absolute population evidence, by correctness).
\begin{tabular}{lrrrrr}
\toprule
Term & $N$ & Estimate & SE & $t$ & $p$ \\
\midrule
Correct-trial slope
& 60{,}532 & $+0.678$ & $0.024$ & $28.35$ & $7.5\times10^{-177}$ \\
Error-trial slope
& 39{,}468 & $-0.589$ & $0.031$ & $-19.06$ & $5.3\times10^{-81}$ \\
Interaction
& 100{,}000 & $+1.267$ & $0.039$ & $32.43$ & $9.0\times10^{-231}$ \\
\bottomrule
\end{tabular}
\end{table}
\begin{table}[t]
\centering
\small
\caption{Gemini Flash 3: contrast task results}
\label{tab:fiercefalcon_contrast_90}
\makebox[\textwidth][c]{%
\begin{tabular}{llll}
\toprule
\multicolumn{4}{l}{\textbf{A. Psychometric fit}} \\
\midrule
Analysis & $N$ & Key estimate & Summary \\
\midrule
Choice psychometric
& 16{,}800
& slope $= 119.8$, PSE $= +0.014$, pseudo-$R^2 = 0.738$
& Steep contrast sensitivity with small positive boundary shift \\
\midrule
\multicolumn{4}{l}{}\\[-0.8em]
\multicolumn{4}{l}{\textbf{B. Fixed stimulus-bin prediction of accuracy}} \\
\midrule
Analysis & $N$ & Key estimate & Summary \\
\midrule
Fixed-bin logistic model
& 9{,}459
& $\beta = 0.343$, OR $= 1.41$, $p = 8.5 \times 10^{-30}$
& $|\mathrm{LD}|$ predicted correctness within stimulus bins \\
\midrule
\multicolumn{4}{l}{}\\[-0.8em]
\multicolumn{4}{l}{\textbf{C. X-pattern slopes}} \\
\midrule
Branch / contrast & $N$ & Standardized slope & Summary \\
\midrule
Correct trials
& 14{,}571
& $+13.84$
& $|\mathrm{LD}|$ increased with stimulus strength \\
Error trials
& 1{,}811
& $-19.10$
& $|\mathrm{LD}|$ decreased with stimulus strength \\
Correct--error interaction
& 16{,}382
& $+32.95$
& Strong X-pattern interaction \\
\bottomrule
\end{tabular}}
\end{table}
\begin{table}[ht]
\centering
\caption{
    CLEVR nested and fixed-blur-bin analyses. Blur strength is defined as negative blur $\sigma$, so larger values indicate less degraded images. The nested model compares blur strength alone with blur strength plus $|\mathrm{LD}|$ as predictors of correctness. The fixed-bin model tests whether $|\mathrm{LD}|$ predicts correctness after controlling for coarse blur bin. Difficulty-controlled columns report the $|\mathrm{LD}|$ coefficient after additionally controlling for CLEVR scene and question difficulty variables. Coefficients marked with $^{***}$ are significant at $p<.001$. For Gemini Flash 3 count, the difficulty-controlled fixed-bin model was not estimable because of collinearity among controls.
}
\label{tab:clevr_nested_fixed_bin_summary}
\resizebox{\textwidth}{!}{
\begin{tabular}{lrrrrrrr}
\toprule
Dataset                            & $N$        & Acc.    & AUROC blur & AUROC blur+$|\mathrm{LD}|$ & $\Delta$AUROC & Fixed-bin $\beta_{|\mathrm{LD}|}$ & Diff.-controlled $\beta_{|\mathrm{LD}|}$ \\
\midrule
Qwen count                         & $34{,}000$ & $0.738$ & $0.666$    & $0.726$                    & $+0.060$      & $1.882^{***}$                     & $1.316^{***}$                             \\
Gemini Flash 3 count, 90\% cut     & $11{,}000$ & $0.734$ & $0.632$    & $0.703$                    & $+0.072$      & $0.162^{***}$                     & $1.210^{\dagger}$                         \\
Qwen existence                     & $28{,}000$ & $0.761$ & $0.774$    & $0.825$                    & $+0.050$      & $0.456^{***}$                     & $1.015^{***}$                             \\
Gemini Flash 3 existence, 90\% cut & $11{,}000$ & $0.721$ & $0.673$    & $0.703$                    & $+0.029$      & $0.106^{***}$                     & $0.693^{***}$                             \\
\bottomrule
\end{tabular}
}

\vspace{1ex}

\begin{flushleft}
\footnotesize
\textit{Note.} Fixed-bin coefficients are from logistic models predicting correctness from $|\mathrm{LD}|$ while controlling for coarse blur bin. Difficulty-controlled coefficients are standardized $|\mathrm{LD}|$ coefficients from nested logistic models controlling for blur strength and CLEVR difficulty variables. $^{\dagger}$The individual Wald test for the Gemini Flash 3 count coefficient was not significant, but the likelihood-ratio test showed that $|\mathrm{LD}|$ significantly improved the model over blur plus difficulty controls ($p=1.68\times10^{-57}$).
\end{flushleft}
\end{table}


\begin{table}[ht]
\centering
\caption{Correct/error branch slopes in the CLEVR X-pattern analysis. Slopes are in units of $|\mathrm{LD}|$ per standard-deviation increase in stimulus strength. The interaction is the correct-branch slope minus the error-branch slope.}
\label{tab:clevr_xpattern_triallevel}
\resizebox{\textwidth}{!}{
\begin{tabular}{lrrrrrrr}
\toprule
Dataset & $N$ & Accuracy & Error slope & Correct slope & Interaction & Interaction $t$ & Interaction $p$ \\
\midrule
Qwen count
& $34{,}000$ & $0.738$
& $+0.059$
& $+0.346$
& $+0.287$
& $59.92$
& $<10^{-300}$ \\

Gemini Flash 3 count, 90\% cut
& $11{,}000$ & $0.734$
& $-0.089$
& $+1.180$
& $+1.269$
& $8.13$
& $<10^{-15}$ \\

Qwen existence
& $28{,}000$ & $0.761$
& $+0.019$
& $+1.218$
& $+1.199$
& $46.73$
& $<10^{-300}$ \\

Gemini Flash 3 existence, 90\% cut
& $11{,}000$ & $0.721$
& $+0.136$
& $+0.623$
& $+0.487$
& $3.82$
& $1.4\times10^{-4}$ \\
\bottomrule
\end{tabular}
}
\end{table}

\end{document}